\pdfoutput=1

\documentclass[11pt]{article}

\usepackage[]{EMNLP2023}

\usepackage{times}
\usepackage{latexsym}

\usepackage[T1]{fontenc}

\usepackage[utf8]{inputenc}

\usepackage{microtype}

\usepackage{inconsolata}

\usepackage{booktabs}
\usepackage{multicol}
\usepackage{graphicx}
\usepackage{tabularx}
\usepackage{hyperref}
\usepackage{listings}
\usepackage{array}
\usepackage{xcolor}
\usepackage{algorithm}
\usepackage{algpseudocode}
\usepackage{enumitem}
\usepackage{multirow}
\usepackage{wrapfig}
\usepackage{mwe}
\usepackage[export]{adjustbox}

\usepackage{caption}
\usepackage{subcaption}

\usepackage{amsmath}

\title{In-Context Example Selection via Similarity Search\\
Improves Low-Resource Machine Translation}

\author{Armel Zebaze 
\quad Benoît Sagot\quad\textbf{Rachel Bawden}\\
Inria, Paris, France\\
\texttt{\{armel.zebaze-dongmo,benoit.sagot,rachel.bawden\}@inria.fr}} 

\begin{document}
\maketitle
\begin{abstract}
The ability of generative large language models (LLMs) to perform in-context learning has given rise to a large body of research into how best to prompt models for various natural language processing tasks.
In this paper, we focus on machine translation (MT), a task that has been shown to benefit from in-context translation examples. However no systematic studies have been published on how best to select examples, and mixed results have been reported on the usefulness of similarity-based selection over random selection.
We provide a study covering multiple LLMs and multiple in-context example retrieval strategies, comparing multilingual sentence embeddings. We cover several language directions, representing different levels of language resourcedness (English into French, German, Swahili and Wolof).
Contrarily to previously published results, we find that sentence embedding similarity can improve MT, especially for low-resource language directions, and discuss the balance between selection pool diversity and quality.
We also highlight potential problems with the evaluation of LLM-based MT and suggest a more appropriate evaluation protocol, adapting the COMET metric to the evaluation of LLMs.
Code and outputs are freely available at \url{https://github.com/ArmelRandy/ICL-MT}.\footnote{We report implementation details in Appendix~\ref{appendix:experimental_details}.}
%

\end{abstract}

\section{Introduction}
In-context learning (ICL, \citet{NEURIPS2020_1457c0d6}) for large language models (LLMs) has proved successful for various tasks, including machine translation (MT) \citep{bawden-yvon-2023-investigating, 10.5555/3618408.3620130, zhu2023multilingual, gpt-mt-2023, xu2024a, lyu-etal-2024-paradigm}. Usually, in-context examples for MT are randomly sampled from a parallel corpus. However, existing work in question answering \citep{liu-etal-2022-makes} and text classification \citep{pmlr-v139-zhao21c} has shown that the choice of in-context examples considerably influences ICL outcomes. This aspect has been explored in MT through example retrieval via similarity search, where in-context examples are chosen based on their similarity to the sentence to be translated. However, consensus on its efficacy has not been reached. \citet{DBLP:conf/acl/VilarFCLRF23} found that retrieving similar sentences does not yield more benefits than selecting them randomly when the selection pool contains only high-quality samples. Their experiments focused on high-resource directions. \citet{zhu2023multilingual} and \citet{gpt-mt-2023} arrived at the same conclusion when examining other high-resource directions. However, \citet{agrawal-etal-2023-context} surpassed the random baseline by using examples retrieved with BM25 and further improved performance through a re-ranking procedure. \citet{10.5555/3618408.3620130} observed a correlation between the use of similar examples and performance but cautioned that the correlation may not be strong enough.
Not only do these mixed results show that it is not clear whether example selection can provide gains, but the impact of few-shot example selection for low-resource languages remains underexplored. Existing research also often overlooks the impact of the size and quality of the selection pool, and there is a lack of analysis across LLMs of different scales.

In this work, we aim to address these gaps by systematically analyzing example retrieval via similarity search. We benchmark multiple similarity metrics based on multilingual sentence embeddings across various open-access LLMs. We consider translations from English to French, German, Swahili and Wolof to account for different levels of resourcedness. We compare the use of sentence embeddings and existing approaches, and we assess the robustness of this strategy against different selection pool compositions when translating from English to Swahili. Additionally, we highlight potential problems with the evaluation of LLM-based MT and propose a more appropriate evaluation protocol. Our analysis suggests that example retrieval via similarity search only marginally improves MT over random sampling for high-resource languages. However, for the first time, we observe significant gains across all metrics when translating into low-resource languages. These results are observable across LLMs of multiple scales.

\begin{figure*}[ht]
    \centering
    \includegraphics[width=0.95\textwidth]{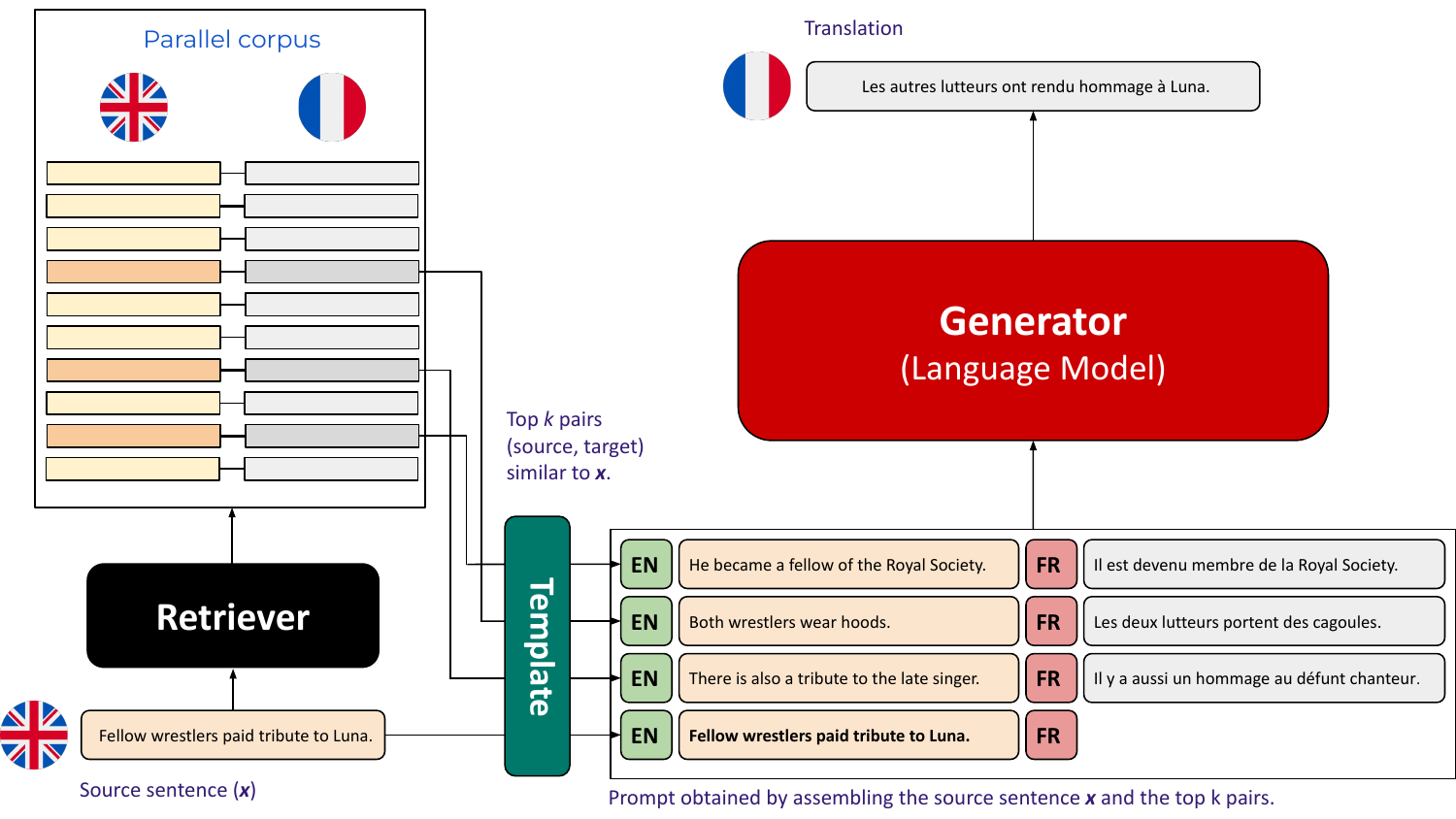}
    \caption{An overview of example retrieval via similarity search for MT. $k$ sentences are first retrieved from the example pool (parallel corpus) based on their similarity to the source sentence. The retrieved sentence pairs are then assembled (as few-shot examples) with the source sentence into a prompt that is fed to a LLM for translation.}
    \label{fig:fig1}
\end{figure*}

\section{Background and Related Work}

\paragraph{In-Context Learning (ICL).} After \citet{NEURIPS2020_1457c0d6} demonstrated GPT-3's strong zero-shot and few-shot abilities on language understanding benchmarks, the research community has put a lot of effort into empirically analyzing ICL. \citet{pmlr-v139-zhao21c} showed that the prompt format, the quality of the examples and their order all have an effect on performance, although it has been shown, for example by \citet{min-etal-2022-rethinking} for few-shot text classification, that performance can plateau as the number of examples included increases. Another line of work explored the design of prompting strategies with most results obtained on reasoning tasks: chain of thought \citep{NEURIPS2022_9d560961, NEURIPS2022_8bb0d291, zhang2023automatic}, self-consistency \citep{wang2023h, chen2023universal} and tree of thoughts \citep{NEURIPS2023_271db992}. 


\paragraph{Using LLMs for Machine Translation.} In MT, comparing LLMs and understanding their behaviour in few-shot settings
has motivated multiple studies.
\citet{lin-etal-2022-shot} showed that XGLM 7.5B outperforms GPT-3 6.7B in 32-shot for multiple translation directions. \citet{DBLP:conf/acl/VilarFCLRF23} used PALM \citep{chowdhery2022palm} for few-shot MT. They ran experiments on high resource languages and concluded that the quality of the selection pool has a high impact on few-shot MT. 
\citet{10.5555/3618408.3620130} and \citet{bawden-yvon-2023-investigating} respectively analyzed GLM-130B \citep{zeng2023glmb} and BLOOM \citep{workshop2023bloom} for few-shot MT. They both highlighted the importance of the prompt format inter alia. \citet{gpt-mt-2023} demonstrated the competitiveness of GPT models prompted in few-shot against commercial MT systems. Most of these works focus on high-resource languages, but \citet{gpt-mt-2023} used two low-resource languages (Hausa and Icelandic) to demonstrate that GPT models lag behind the best MT systems and \citet{bawden-yvon-2023-investigating} studied 1-shot MT between low-resource languages pairs. \citet{zhu2023multilingual} conducted a systematic study in which they compared eight LLMs for few-shot MT in 102 languages covering different resource levels, although most of their experiments were done with eight randomly picked few-shot examples.


\paragraph{Similarity Search for Example Selection.} While a majority of works, including those in MT, use few-shot examples that are randomly selected, others explore how selecting particular examples can impact performance. This is often achieved by mining sentences similar to the one to be processed, generally based on sentence vector representations based on token-level language models (e.g.~RoBERTa, \citealp{liu2019roberta}) or on sentence embedding models (e.g.~LASER2, \citealp{heffernan-etal-2022-bitext}). \citet{liu-etal-2022-makes} showed that $k$-NN retrieval with fine-tuned RoBERTa models improved GPT-3 performance on question answering and table-to-text generation tasks. \citet{DBLP:conf/acl/VilarFCLRF23} implemented $k$-NN retrieval with RoBERTa and bag-of-word embeddings for few-shot MT between high-resource language pairs. Similarly, \citet{zhu2023multilingual} compared BM25 \citep{robertson1995okapi}  to example retrieval with a sentence embedding for MT from English to German and Russian. They both conclude that the use of similar examples is comparable to that of random examples for a high quality selection pool. \citet{gpt-mt-2023} used LaBSE \citep{feng-etal-2022-language} to build a high-quality selection pool and/or to perform high-quality example selection. Their experiments on German, Russian and Chinese showed the irrelevance of quality selection from a high quality selection pool. \citet{10.5555/3618408.3620130} studied the correlation between shot selection and MT performance for multiple strategies 
including example retrieval with LASER2. Their work mostly focused on Chinese and German for which they reported mixed results, 
and \citet{agrawal-etal-2023-context} explored example selection with BM25 and showed that their re-ranking procedure could improve BLEU scores. The variability in the conclusions regarding the efficacy of similarity-based selection methods highlights the necessity for a more systematic study  covering both high-resource languages  and low-resource languages, which are frequently excluded from these experiments.

\section{Example Retrieval via Similarity Search}

Example retrieval via similarity search is a selection strategy for ICL. The idea is to use the input in order to retrieve similar (input, output) pairs from a pool of labeled data, which can then be used as few-shot examples (see Figure~\ref{fig:fig1}). It revolves around the following parameters:

\begin{enumerate}[leftmargin=*]
    \item \textbf{A pool $\mathcal{P}$ from which to retrieve examples for the source sentence $x$}. For MT, the pool corresponds to a set of parallel sentence pairs.
    \item \textbf{The number $k$ of few-shot examples to retrieve from $\mathcal{P}$}. By definition, $k \leq |\mathcal{P}|$.
    \item \textbf{A retriever} $\mathcal{R}$. In a similar spirit to RAG \citep{NEURIPS2020_6b493230}, its role is to identify similar example pairs to add to the context in\if order to build\fi{} the input prompt. This similarity can be syntactic or semantic depending on the aspects of the sentence we decide to analyze. In this work, we model similarity with cosine similarity and we compare this to $n$-gram metrics. 
    \item \textbf{A template to format each example}. This is used to assemble the sentence to translate and the few-shot examples to construct the prompt to be fed to the LLM. By default, the most similar demonstration is the closest to the sentence to translate. We ablate this choice in Appendix~\ref{appendix:reverse}.
    \item \textbf{An LLM}. The LLM ($p_\theta$) is fed with the prompt in order to obtain the translation. We test a variety of decoder-based LLMs in our study. 
\end{enumerate}

In MT, $\mathcal{P}$ consists of the source and target sides of parallel data. Retrieval can be done by analyzing the similarity of the sentence to translate to either the source or target side of each pair in $\mathcal{P}$. This implies that there are two possible approaches to example retrieval, which we refer to as \textit{source-to-source} and \textit{source-to-target}. By default (and unless specified otherwise) we use the \textit{source-to-source} retrieval approach (See Appendix~\ref{appendix:section-s2t} for the \textit{source-to-target} approach).

\section{Experimental Setup}

\paragraph{Datasets}\label{datasets}
We work on MT from English (eng) as it is more challenging than translating into English.\footnote{See Appendix~\ref{appendix:to-english} for translation into English.} and choose to work with four target languages: two high-resource, French (fra) and German (deu), one mid-resource, Swahili (swa) and one low-resource, Wolof (wol). 
For evaluation, we use the FLORES-200 \citep{goyal-etal-2022-flores, nllb2022} devtest set containing 1012 examples. We use the FLORES-200 dev set (997 examples) as the selection pool $\mathcal{P}$. We also consider 20,000 examples from the NLLB dataset{} \citep{nllb2022}  for experiments involving pool extension. We refer to this additional dataset as $\mathcal{U}$.

\paragraph{Retrievers} \label{retrievers}

We compare five multilingual sentence embeddings: SONAR \citep{Duquenne:2023:sonar_arxiv}, Embed v3,\footnote{\url{https://txt.cohere.com/introducing-embed-v3/}} E5 \citep{wang2022text}, LaBSE
\citep{feng-etal-2022-language} 
and LASER2
\citep{heffernan-etal-2022-bitext}. 
We compare against the following approaches: BM25{} \citep{robertson1995okapi}, R-BM25 (consisting in retrieving the top 100 similar candidates with BM25, re-ranking them using the algorithm outlined in \citep{agrawal-etal-2023-context} and choosing the $k$ first for ICL), BLEU \citep{10.3115/1073083.1073135} and RoBERTa \citep{liu2019roberta} embeddings.\footnote{More precisely, we use the last hidden state of the first token and send it to the pooling layer. We use the RoBERTa-large model.} We also compare against a baseline where the $k$ in-context examples are randomly sampled from the pool, reporting the average score over three different seeds.

\paragraph{Models}

We test multiple LLMs in our experiments. For reproducibility, we consider state-of-the-art open-access LLMs: 
BLOOM 7B \citep{workshop2023bloom}, 
OLMo~7B \citep{groeneveld2024olmo}, Gemma (2B, 7B) \citep{gemmateam2024gemma} LLaMA-2 (7B, 13B and 70B) \citep{touvron2023llama}, Mistral~7B~v0.1 \citep{jiang2023mistral} and Mixtral~8x7B~v0.1~\citep{jiang2024mixtral}. 

\paragraph{Evaluation metrics}
Historically, BLEU \citep{10.3115/1073083.1073135} has been the standard MT evaluation metric. The recent advances in deep learning fueled the emergence of neural metrics, one of the most successful being COMET \citep{rei-etal-2020-comet}, which is better correlated with human judgements than BLEU 
\citep{rei-etal-2022-comet}.
Despite this superiority, 
COMET has some limitations for evaluating MT by LLMs. 
First, it is inherently limited by the language coverage of its encoder, impairing its reliability for unseen languages (e.g.~Wolof). Moreover, it is not robust to the issues of translation in the wrong 
language and empty translations. These issues were previously taken for granted when designing metrics, since it was always assumed that MT systems were designed to produce text in the correct language. However, they have become relevant with the use of LLMs for MT, since these models are not trained for MT specifically, and therefore the premise of a translation being in the correct language does not always hold. The two problems are more likely to appear in zero-shot settings and when few in-context examples are used, especially when prompting a model to generate a low-resource language. We propose to alleviate them with a simple correction protocol consisting in setting the score of a translation to 0 if it is either empty or written in the wrong target language. We name this variant Language-Aware COMET (laCOMET) which preserves the benefits of COMET while making it robust to the previously mentioned issues. It is worth noting that laCOMET is strictly equivalent to COMET for sentences that do not exhibit the issues that motivated its creation (i.e.~non-empty translations in the correct language). 

We use laCOMET, based on COMET 22{} \citep{rei-etal-2022-comet} as our main metric. We use fasttext
\citep{bojanowski-etal-2017-enriching, nllb2022} for language identification, which supports more than 200 languages including those we work with. For transparency, we also include BLEU calculated using SacreBLEU \citep{post-2018-call}\footnote{nrefs:1$\vert$case:mixed$\vert$eff:no$\vert$tok:flores200$\vert$smooth:exp$\vert$\\version:2.3.2} and COMET in the appendix.

\section{Experiments}

We begin by exploring template selection (Section~\ref{prompt-selection}) in order to select the template we will use for the remainder of the experiments. In Section~\ref{se-benchmarking} we do a systematic study of example retrieval with several multilingual sentence embeddings for different numbers of in-context examples and families of LLMs, and in Section ~\ref{comparison} we compare example retrieval with the best performing sentence embedding and the previously mentioned alternative approaches. In Section~\ref{robustness} we study the robustness of example retrieval to the size and the diversity of the pool of examples. Finally, in Section~\ref{scalability}, we focus on English to Swahili and analyze example retrieval for various LLMs at different scales.

\subsection{Template selection} \label{prompt-selection}

We carry out a preliminary investigation to choose a strong template for our subsequent MT experiments. We compare six potential MT templates (listed in Table~\ref{tab:prompt-table}) in 0-shot and 5-shot settings for three models and the four directions. The BLEU scores are shown in Table~\ref{tab:prompt-result}\footnote{We choose to report initial BLEU scores for the different prompts rather than laCOMET scores (the main metric used in the rest of the paper), as  BLEU scores are informative for MT specialists in terms of getting intuitions about absolute MT quality, and the score differences we observe between prompts are sufficiently great to be captured by BLEU.}. The best template for a model does not necessarily work well with another model in the zero-shot setting (e.g.~T3 $\geq$ T5 for LLaMA~2~7B but not for Mistral~7B~v0.1). We notice that having the end of the prompt written in the target language can dramatically improve zero-shot MT; using template T2 instead of template T1 gives an absolute gain of 11.5 BLEU for BLOOM~7B1, 5.5 for Mistral~7B~v0.1 and 0.8 for LLaMA~2~7B for eng$\rightarrow$fra. For eng$\rightarrow$deu, T2 surpasses T1 by 0.2 BLEU for BLOOM~7B1, 4.4 for Mistral~7B~v0.1 and 2.7 for LLaMA~2~7B. Similarly, significant gains are observed when using T4 instead of T3. We hypothesize that these improvements are attributed to the fact that the prompt ending in the target language encourages the model to continue generation in that language, reducing the occurrence of unrelated outputs. The presence of a colon ($\textbf{:}$) at the end of the prompt can have a negative effect on some LLMs such as Mistral~7B~v0.1 and LLaMA~2~7B, making them generate dates (with the format $\textsc{YYYY-MM-DD}$). The performance disparities among templates T1, T2, T5 and T6 disappear in the 5-shot setting but the negative impact of the colon keeps templates T3 and T4 behind. Translating into low-resource languages gives poor scores in the zero-shot setting, which prevents a reliable comparison of the templates. However, the scores are generally close to each other. T1, T2, T5, and T6 are the optimal templates for eng$\rightarrow$swh and eng$\rightarrow$wol in few-shot scenarios for all three LLMs. The summary of this analysis is that zero-shot performance varies greatly across templates as observed by \citep{10.5555/3618408.3620130}. This discrepancy tends to disappear in few-shot except for adversarial templates. Any template between T1, T2, T5 and T6 would allow a fair comparison between models in few-shot scenarios. In the rest of this work, we choose to use template T5 because of its simplicity and good few-shot performance.

\begin{table*}[ht]
\centering\small
\begin{tabular}{lll}
\toprule
\multicolumn{1}{c}{\bf ID}  &\multicolumn{1}{c}{\bf Template} & \multicolumn{1}{c}{\bf Example (eng$\rightarrow$fra)}\\
\midrule
T1         & \textcolor{black}{[src]} $\diamond$ [source] $\diamond$ translates into $\diamond$ \textcolor{black}{[tgt]} $\diamond$ & English $\diamond$ I live in Paris. $\diamond$ translates into $\diamond$ French $\diamond$\\
T2         &  [src]\textsubscript{\bf src} $\diamond$ [source] $\diamond$ translates into $\diamond$ [tgt]\textsubscript{\bf tgt} $\diamond$ & English $\diamond$ I live in Paris. $\diamond$ translates into $\diamond$ Français $\diamond$\\
T3         & \textcolor{black}{[src]}: [source] $\diamond$ \textcolor{black}{[tgt]}: & English: I live in Paris. $\diamond$ French: \\
T4         & [src]\textsubscript{\bf src}: [source] $\diamond$ [tgt]\textsubscript{\bf tgt}: & English: I live in Paris. $\diamond$ Français: \\
T5         & \textcolor{black}{[src~sentence]} $\diamond$ [source] $\diamond$ \textcolor{black}{[tgt~translation]} $\diamond$ & English sentence $\diamond$ I live in Paris. $\diamond$ French translation $\diamond$\\
T6         & [src~sentence]\textsubscript{\bf src} $\diamond$ [source] $\diamond$ [tgt~translation]\textsubscript{\bf tgt} $\diamond$ & English sentence $\diamond$ I live in Paris. $\diamond$ Traduction en français $\diamond$\\
\bottomrule
\end{tabular}
\caption{Templates considered for template selection. \textit{src} represents the source language (e.g.~English), \textit{tgt} the target language (e.g.~French) and \textit{source} the sentence to translate. The presence of the subscripts \textbf{src} and \textbf{tgt} indicates that the words are written in the source language and the target language, respectively.}
\label{tab:prompt-table}
\end{table*}

\begin{table*}[ht]
\centering\small
\begin{tabular}{lrrrrrrrrrrrrr}
\toprule
\multicolumn{1}{c}{\bf }  &\multicolumn{6}{c}{\bf 0-shot} & \multicolumn{1}{c}{\bf } &\multicolumn{6}{c}{\bf 5-shot}\\
{} & T1 & T2 & T3 & T4 & T5 & T6 & & T1 & T2 & T3 & T4 & T5 & T6 \\
\midrule
\multicolumn{2}{l}{BLOOM~7B1} \\
\midrule
eng$\rightarrow$fra  & 2.6 & 14.1 & 10.5 & 22.8 & 27.5 & \bf 41.7 & & 46.6 & 46.9 & 46.4 & 46.6 & 46.7 & \bf 47.0\\
eng$\rightarrow$deu  & 2.1 & 2.3 & 3.1 & 6.4 & \bf 6.6 & 1.3 & & 14.0 & 14.0 & 13.5 & 13.8 & 13.9 & \bf 14.1\\
eng$\rightarrow$swh  & 1.4 & 1.7 & 1.5 & 1.6 & 3.2 & \bf 3.9 & & \bf 10.8 & 10.7 & 10.5 & 10.4 & 10.5 & 10.2\\
eng$\rightarrow$wol  & 1.3 & 1.3 & 1.7 & 1.7 & \bf 2.5 & 0.5 & & 1.5 & 1.5 & 1.4 & 1.4 & 1.6 & \bf 1.8\\
\midrule
\multicolumn{2}{l}{Mistral~7B~v0.1} \\
\midrule
eng$\rightarrow$fra  & 8.9 & 14.4 & 26.4 & 24.2 & \bf 44.6 & 40.8 & & \bf 48.3 & 48.1 & 47.0 & 46.8 & 48.0 & 48.1\\
eng$\rightarrow$deu  & 7.8 & 12.2 & 14.6 & 16.5 & \bf 33.0 & 31.7 & & 37.4 & \bf 37.6 & 35.2 & 35.2 & 37.3 & 37.3\\
eng$\rightarrow$swh  & \bf 2.8 & 2.7 & 1.3 & 1.5 & 2.4 & 2.7 & & 2.7 & 2.8 & 2.8 & \bf 2.9 & 2.8 & 2.8\\
eng$\rightarrow$wol  & \bf 2.8 & 2.8 & 0.2 & 0.2 & 2.6 & 0.7 & & 2.2 & 2.2 & 1.8 & 1.7 & \bf 2.3 & 2.1\\
\midrule
\multicolumn{2}{l}{LLaMA~2~7B} \\
\midrule
eng$\rightarrow$fra  & 10.2 & 11.0 & 19.3 & \bf 28.2 & 5.3 & 8.4 & & \bf 45.4 & 45.3 & 41.3 & 41.3 & 45.2 & 45.3\\
eng$\rightarrow$deu  & 9.8 & 12.5 & 15.1 & \bf 19.4 & 5.1 & 3.8 & & 35.2 & 35.2 & 30.0 & 31.1 & \bf 35.2 & 34.9\\
eng$\rightarrow$swh  & 1.1 & 1.3 & 1.0 & 0.9 & \bf 1.3 & 0.9 & & 2.7 & \bf 2.8 & 1.6 & 0.7 & \bf 2.8 & 2.7\\
eng$\rightarrow$wol  & \bf 1.5 & 1.5 & 0.0 & 0.0 & 0.2 & 0.2 & & 2.1 & 2.1 & 1.5 & 1.5 & 2.1 & \bf 2.2\\
\bottomrule
\end{tabular}
\caption{Comparison of BLEU scores on the FLORES-200 devtest set with three LLMs and the six templates (T1--T6) detailed in Table~\ref{tab:prompt-table} for 0-shot and 5-shot settings. 5-shot examples are sampled uniformly at random. We report the average BLEU score across three runs with different seeds.}
\label{tab:prompt-result}
\end{table*}

\subsection{Benchmarking of example retrieval with multilingual sentence embeddings} \label{se-benchmarking}

\begin{table*}[t]
\tiny
\begin{center}
\resizebox{1.0\textwidth}{!}{
\begin{tabular}{llrrrrrrrrrrrr}
\toprule
{\bf Model}& {\bf Method}  & \multicolumn{3}{c}{\bf eng$\rightarrow$fra} & \multicolumn{3}{c}{\bf eng$\rightarrow$deu} & \multicolumn{3}{c}{\bf eng$\rightarrow$swh} & \multicolumn{3}{c}{\bf eng$\rightarrow$wol}\\
& {} & 1 & 5 & 10 & 1 & 5 & 10 & 1 & 5 & 10 & 1 & 5 & 10\\
\midrule
BLOOM~7B1&Embed v3 & 79.6 & 86.7 & \textbf{86.7} & 55.2 & \bf 60.1 & \textbf{61.0} & 58.6 & \textbf{68.4} & 69.4 & 50.4 & 50.2 & 50.7\\
&E5 & \textbf{80.4} & 86.6 & \bf 86.7 & 54.5 & \bf 60.1 & 60.6 & \textbf{59.8} & 68.2 & 69.3 & \textbf{50.9} & \textbf{51.4} & 50.7 \\
&LaBSE & 79.4 & 86.7 & \bf 86.7 & 55.1 & 59.9 & 60.5 & 58.3 & 67.8 & 69.2 & 49.9 & 51.2 & \textbf{52.3}\\
&LASER2 & 79.2 & 86.6 & \bf 86.7 & 55.1 & 59.9 & 59.6 & 58.0 & 67.7 & 67.8 & 48.5 & 50.1 & 50.9 \\
&SONAR & 79.8 & \textbf{86.8} & 86.6 & \textbf{55.3} & \textbf{60.1} & 60.8 & 57.4 & 68.3 & \textbf{69.6} & 50.2 & 50.4 & 51.6 \\
&Random & 77.3 & 86.5 & 86.6 & 52.8 & 57.7 & 57.7 & 56.9 & 65.1 & 66.0 & 46.5 & 45.1 & 46.4 \\
\midrule
Mistral~7B~v0.1&Embed v3 & \textbf{86.2} & \textbf{87.0} & \textbf{87.0} & 83.5 & 85.7 & \bf 85.9 & \textbf{37.5} & \textbf{41.4} & 43.3 & 36.5 & 44.1 & 44.7\\
&E5 & 85.7 & \bf 87.0 & 86.9 & 83.4 & 85.2 & 85.5 & 37.3 & 41.3 & 43.2 & 36.6 & 44.3 & 44.4 \\
&LaBSE & 86.2 & 86.7 & \bf 87.0 & 83.3 & 85.3 & 85.6 & 37.0 & 40.1 & 42.3 & \textbf{36.7} & 42.6 & 44.6\\
&LASER2 & 86.1 & 86.9 & \bf 87.0 & 83.5 & 85.6 & 85.5 & 35.3 & 38.0 & 40.3 & 32.0 & 42.1 & 43.3 \\
&SONAR & 86.1 & 86.9 & \bf 87.0 & \textbf{83.6} & \textbf{85.8} & \textbf{85.9} & 37.2 & 40.6 & \textbf{43.5} & 36.4 & \textbf{45.0} & \textbf{46.1} \\
&Random & 85.8 & 86.5 & 86.6 & 83.0 & 85.4 & 85.5 & 32.7 & 33.5 & 33.8 & 26.7 & 33.2 & 36.0 \\
\midrule
LLaMA~2~7B&Embed v3 & 85.8 & 86.1 & 86.3 & 84.0 & 84.9 & 85.0 & \textbf{45.7} & \textbf{43.7} & \textbf{45.6} & 41.8 & 46.2 & \textbf{47.1}\\
&E5 & 85.8 & \textbf{86.2} & \textbf{86.4} & \textbf{84.1} & 85.2 & 85.2 & 45.1 & 43.3 & 45.3 & \textbf{42.3} & \textbf{46.5} & 46.9 \\
&LaBSE & 85.6 & 86.0 & 86.2 & \bf 84.1 & 85.1 & 85.1 & 44.2 & 42.5 & 44.7 & 40.0 & 43.7 & 45.6\\
&LASER2 & 85.8 & \bf 86.2 & 86.2 & 83.6 & 85.0 & 85.2 & 41.2 & 40.1 & 42.1 & 38.7 & 42.5 & 43.3 \\
&SONAR & \textbf{85.9} & 86.1 & 86.3 & 83.8 & \textbf{85.3} & \textbf{85.4} & 45.2 & 43.2 & 45.5 & 39.7 & 45.9 & 46.7 \\
&Random & 85.6 & 85.9 & 86.0 & 83.6 & 84.8 & 85.0 & 35.4 & 34.7 & 35.8 & 34.4 & 34.7 & 36.5 \\
\midrule
Gemma~7B&Embed v3 & 87.5 & \bf 88.0 & \bf 88.1 & 86.7 & 87.3 & 87.5 & 79.0 & 80.7 & \textbf{81.4} & 39.0 & 45.2 & 48.0\\
&E5 & 87.4 & 87.9 & \bf 88.1 & 86.9 & 87.4 & \bf 87.6 & \bf 79.4 & 80.5 & 81.2 & \textbf{39.5} & 45.0 & \textbf{48.4} \\
&LaBSE & \bf 87.7 & 87.9 & 88.0 & \textbf{87.1} & \textbf{87.6} & 87.3 & 79.1 & \textbf{80.8} & 81.1 & 37.0 & 44.4 & 47.8\\
&LASER2 & 87.5 & 87.9 & 87.9 & \bf 87.1 & 87.3 & 87.2 & \textbf{79.4} & 80.6 & 80.5 & 36.0 & 43.9 & 47.6 \\
&SONAR & 87.4 & \textbf{88.0} & \textbf{88.1} & 86.8 & \bf 87.6 & \textbf{87.6} & 79.2 & 80.4 & 80.7 & 38.1 & \textbf{45.6} & 48.3 \\
&Random & 87.5 & 87.9 & 88.0 & 86.6 & 87.2 & 87.3 & 78.4 & 79.6 & 79.8 & 30.9 & 37.4 & 40.5 \\
\bottomrule
\end{tabular}
}
\end{center}
\caption{laCOMET results of example retrieval with different sentence embedding methods for $k$-shot settings ($k \in \{1, 5, 10\}$). The best score for each direction is shown in bold.}
\label{tab:lacomet-se}
\end{table*}

We conduct a benchmarking analysis of example retrieval using multilingual sentence embeddings to evaluate their performance and compare them to random sampling\footnote{We provide an analysis of the overlap between their choices in Appendix~\ref{appendix:overlap}.}. As demonstrated in Table~\ref{tab:lacomet-se}, example retrieval with sentence embeddings consistently outperforms random sampling in few-shot scenarios (up to 10-shot). The performance gain is modest when translating into French and German, typically ranging between 0.1 and 0.5 laCOMET for most LLMs we evaluated, and it tends to narrow as the number of in-context examples increases. However, we note a substantial improvement of around 2.5 in German with BLOOM~7B1. We attribute this greater improvement to the relatively poor performance of BLOOM~7B1 in German as German was not officially included in its training data. For translation into Swahili, the use of sentence embeddings yields gains ranging between 1.7 and 3.4 laCOMET for BLOOM~7B1, 0.6 and 1.6 for Gemma~7B. These gains explode and reach 10 laCOMET when translating into Swahili or Wolof with Mistral~7B~v0.1 and LLaMA~2~7B. Furthermore, all sentence embeddings outperform random sampling in a majority of cases. Although there is not a highly significant variation in performance among them, SONAR, Embed v3 and E5 perform slightly better than LaBSE and LASER2 for example retrieval. SONAR yields the best performance with a little advance on Embed~v3 and E5. In summary, the use of similar in-context examples yields modest gains for high-resource languages, consistent with previous findings \citep{10.5555/3618408.3620130}, but we see significant benefits for low-resource languages. We document the same findings in terms of BLEU and COMET in Appendix~\ref{appendix:bleu-comet} and with more LLMs in Appendix~\ref{appendix:additional-results}.

\subsection{Comparing to other approaches} \label{comparison}

We compare the best performing multilingual sentence embeddings model, SONAR against other approaches from the literature in few-shot scenarios. laCOMET scores are given in Table~\ref{tab:lacomet-comparison}\footnote{We report additional results with more LLMs in Appendix~\ref{appendix:additional-results}.}. SONAR demonstrates larger performance gains across all language directions and LLMs. Following SONAR, BM25 emerges as the second-best approach. Its reliance on $n$-gram-(word-)matching inherently positions it as a strong contender for example selection. However, applying the re-ranking proposed by \citet{agrawal-etal-2023-context} fails to further improve BM25 in our experimental setup. We attribute this failure to a lack of diversity in the example pool, which hinders its ability to cover each word of the sentences to translate. While RoBERTa can achieve performance levels comparable to those of SONAR in French and German, it consistently lags behind in Swahili and Wolof. This discrepancy may be attributed to the fact that RoBERTa is not explicitly trained to output similar vector representations for two similar sentences, resulting in worse choices than SONAR. Nevertheless, RoBERTa still outperforms random sampling in our evaluations.

\begin{table*}[t]
\tiny
\begin{center}
\resizebox{1.0\textwidth}{!}{
\begin{tabular}{llrrrrrrrrrrrrr}
\toprule
\multicolumn{1}{c}{\bf Model}&\multicolumn{1}{c}{\bf Metric}  & \multicolumn{3}{c}{\bf eng$\rightarrow$fra} & \multicolumn{3}{c}{\bf eng$\rightarrow$deu} & \multicolumn{3}{c}{\bf eng$\rightarrow$swh} & \multicolumn{3}{c}{\bf eng$\rightarrow$wol}\\
& {} & 1 & 5 & 10 & 1 & 5 & 10 & 1 & 5 & 10 & 1 & 5 & 10\\
\midrule
BLOOM~7B1& SONAR & 79.8 & \textbf{86.8} & 86.6 & \textbf{55.3} & \textbf{60.1} & \textbf{60.8} & 57.4 & \textbf{68.3} & \textbf{69.6} & \textbf{50.2} & \textbf{50.4} & \textbf{51.6} \\
&BM25 & 78.8 & 86.6 & 86.7 & 54.2 & 59.7 & 59.7 & 57.0 & 66.8 & 68.5 & 49.4 & 49.1 & 50.4 \\
&R-BM25 & \textbf{82.0} & 86.4 & 86.5 & 52.9 & 57.7 & 58.6 & 54.8 & 64.3 & 65.3 & 42.4 & 43.8 & 45.8 \\
&BLEU & 78.2 & 86.7 & 86.6 & 53.6 & 59.2 & 59.9 & 57.0 & 66.2 & 67.4 & 49.5 & 49.5 & 50.9 \\
&RoBERTa & 78.5 & 86.7 & \textbf{86.8} & 54.1 & 59.3 & 58.4 & \bf 57.9 & 66.0 & 67.1 & 50.0 & 49.4 & 49.9 \\
&Random & 77.3 & 86.5 & 86.6 & 52.8 & 57.7 & 57.7 & 56.9 & 65.1 & 66.0 & 46.5 & 45.1 & 46.4 \\
\midrule
Mistral~7B~v0.1&SONAR & 86.1 & \textbf{86.9} & \textbf{87.0} & \textbf{83.6} & \textbf{85.8} & \textbf{85.9} & \textbf{37.2} & \textbf{40.6} & \textbf{43.5} & \textbf{36.4} & \textbf{45.0} & \textbf{46.1} \\
&BM25 & \textbf{86.2} & 86.8 & 86.9 & \bf 83.6 & 85.4 & 85.7 & 34.9 & 38.8 & 41.4 & 33.0 & 40.7 & 43.3 \\
&R-BM25 & \bf 86.2 & 86.5 & 86.6 & 83.5 & 85.5 & 85.4 & 31.9 & 33.8 & 34.5 & 24.1 & 28.5 & 32.3 \\
&BLEU & \bf 86.2 & \bf 86.9 & 86.9 & 83.3 & 85.4 & 85.8 & 35.4 & 37.2 & 39.1 & 32.7 & 40.0 & 42.6 \\
&RoBERTa & 85.9 & \bf 86.9 & 86.8 & \bf 83.6 & 85.4 & \bf 85.9 & 33.7 & 35.6 & 37.3 & 32.0 & 39.4 & 42.0 \\
&Random & 85.8 & 86.5 & 86.6 & 83.0 & 85.4 & 85.5 & 32.7 & 33.5 & 33.8 & 26.7 & 33.2 & 36.0 \\
\midrule
LLaMA~2~7B&SONAR & \textbf{85.9} & 86.1 & \textbf{86.3} & \textbf{83.8} & \textbf{85.3} & \textbf{85.4} & \textbf{45.2} & \textbf{43.2} & \textbf{45.5} & \textbf{39.7} & \textbf{45.9} & \textbf{46.7}\\
&BM25 & 85.6 & 86.1 & 86.2 & 83.3 & 84.9 & 85.1 & 40.7 & 40.1 & 42.6 & 38.1 & 43.0 & 45.1 \\
&R-BM25 & 85.5 & 86.0 & 85.8 & 83.1 & 85.0 & 85.0 & 33.5 & 34.2 & 34.8 & 25.4 & 27.7 & 33.1 \\
&BLEU & 85.6 & 86.0 & 86.1 & \bf 83.8 & 85.0 & 85.0 & 38.8 & 39.0 & 40.1 & 36.6 & 41.6 & 43.6 \\
&RoBERTa & 85.6 & \textbf{86.2} & 86.0 & \bf 83.8 & 85.0 & 85.3 & 39.9 & 38.1 & 39.7 & 38.7 & 42.1 & 43.8 \\
&Random & 85.6 & 85.9 & 86.0 & 83.6 & 84.8 & 85.0 & 35.4 & 34.7 & 35.8 & 34.4 & 34.7 & 36.5 \\
\midrule
Gemma~7B&SONAR & 87.4 & 88.0 & \textbf{88.1} & 86.8 & \textbf{87.6} & \textbf{87.6} & \textbf{79.2} & \textbf{80.4} & 80.7 & \textbf{38.1} & \textbf{45.6} & \textbf{48.3} \\
&BM25 & 87.6 & 88.0 & 87.7 & 86.8 & 87.2 & 87.0 & \bf 79.2 & 80.3 & \textbf{80.9} & 35.8 & 43.6 & 47.1 \\
&R-BM25 & 87.6 & 87.9 & 87.7 & 86.8 & 87.1 & 86.8 & 78.3 & 79.7 & 79.6 & 28.2 & 36.2 & 39.1 \\
&BLEU & \textbf{87.7} & 87.9 & \bf 88.1 & \textbf{87.0} & 87.4 & 87.4 & 78.9 & \bf 80.4 & 80.2 & 34.7 & 42.0 & 45.5 \\
&RoBERTa & 87.4 & \textbf{88.1} & \bf 88.1 & 86.7 & 87.3 & 87.4 & 78.8 & 80.2 & 80.1 & 35.6 & 40.6 & 44.0 \\
&Random & 87.5 & 87.9 & 88.0 & 86.6 & 87.2 & 87.3 & 78.4 & 79.6 & 79.8 & 30.9 & 37.4 & 40.5 \\
\bottomrule
\end{tabular}
}
\end{center}
\caption{Comparison of example retrieval with SONAR to baseline methods for $k$-shot settings ($k \in \{1, 5, 10\}$). The best performance (laCOMET) for each direction is shown in bold.}
\label{tab:lacomet-comparison}
\end{table*}

\subsection{Robustness to the quality and the diversity of the selection pool} \label{robustness}

\begin{table*}[ht]
\centering
\small
\begin{tabular}{lrrrrrrrr}
\toprule
 & $\mathcal{P}_1$ & $\mathcal{P}_2$ & $\mathcal{P}_3$ & $\mathcal{P}_4$ & $\mathcal{P}_5$ & $\mathcal{P}_6$ & $\mathcal{P}_7$ & $\mathcal{P}_8$ \\
\midrule
\#FLORES samples ($N_1$) & 10 & 100 & 500 & 997 & 997 & 997 & 997 & 997 \\
\#NLLB samples ($N_2$) & 0 & 0 & 0 & 0 & 1000 & 5000 & 10000 & 20000 \\
\midrule
Vendi Score  & 9.4 & 81.2 & 274.8 & 388.2 & 384.4 & 349.9 & 347.5 & 349.5\\
Perplexity  & 131.0 & 90.9 & 79.9 & 77.4 & 222.7 & 301.8 & 306.3 & 356.5\\
\midrule
BM25 scores & 1.51 & 6.43 & 10.3 & 11.91 & 12.85 & 12.31 & 13.30 & 14.43 \\
SONAR scores & 0.04 & 0.12 & 0.18 & 0.20 & 0.21 & 0.22 & 0.23 & 0.24 \\
\bottomrule
\end{tabular}
\caption{Average of the average similarity between each sentence to be translated and its 10 retrieved examples with SONAR and BM25 for each pool composition.}
\label{tab:average-similarity}
\end{table*}

The performance of ICL is heavily dependent on the diversity and quality of the selection pool. The initial selection pool is a small set of high quality professional translations. Similar to previous works, we extensively studied example retrieval with a high quality pool. In this set of experiments, we compare the behavior of example retrieval with SONAR and BM25 when translating into Swahili across eight different pool compositions $\mathcal{P}_1, \ldots, \mathcal{P}_8$. Each composition includes samples from FLORES-200 dev set and/or samples from the NLLB dataset (see Section~\ref{datasets}). We assess the quality and diversity of each of the eight pool compositions in Table~\ref{tab:average-similarity} with two key metrics: the Vendi Score~\citep{dan2023vendi} and the average perplexity. The Vendi Score, computed with SONAR embeddings, measures diversity, with higher values indicating greater diversity within the composition. The average perplexity, computed using Gemma~2B, measures sample quality, with lower values indicating higher quality samples. In Figure~\ref{fig:compare}, we observe a gradual performance improvement with SONAR and BM25 as the selection pool contains more and more high-quality samples (from $\mathcal{P}_1$ to $\mathcal{P}_4$) in the 5 and 10-shot settings. Although the difference with random sampling is initially modest for both strategies (at $\mathcal{P}_1$), it steadily widens until $\mathcal{P}_4$. The introduction of NLLB samples in the selection pool, which are inherently of lower quality compared to FLORES-200's, induces a decay in the overall quality of outputs for all strategies with random sampling being particularly affected. SONAR emerges as the most robust strategy because it exhibits a lesser performance drop. 
This motivates the use of example selection via similarity search in scenarios where the quality of the pool is heterogeneous or partially known.

\begin{figure*}[ht]
     \centering
     \includegraphics[width=\textwidth]{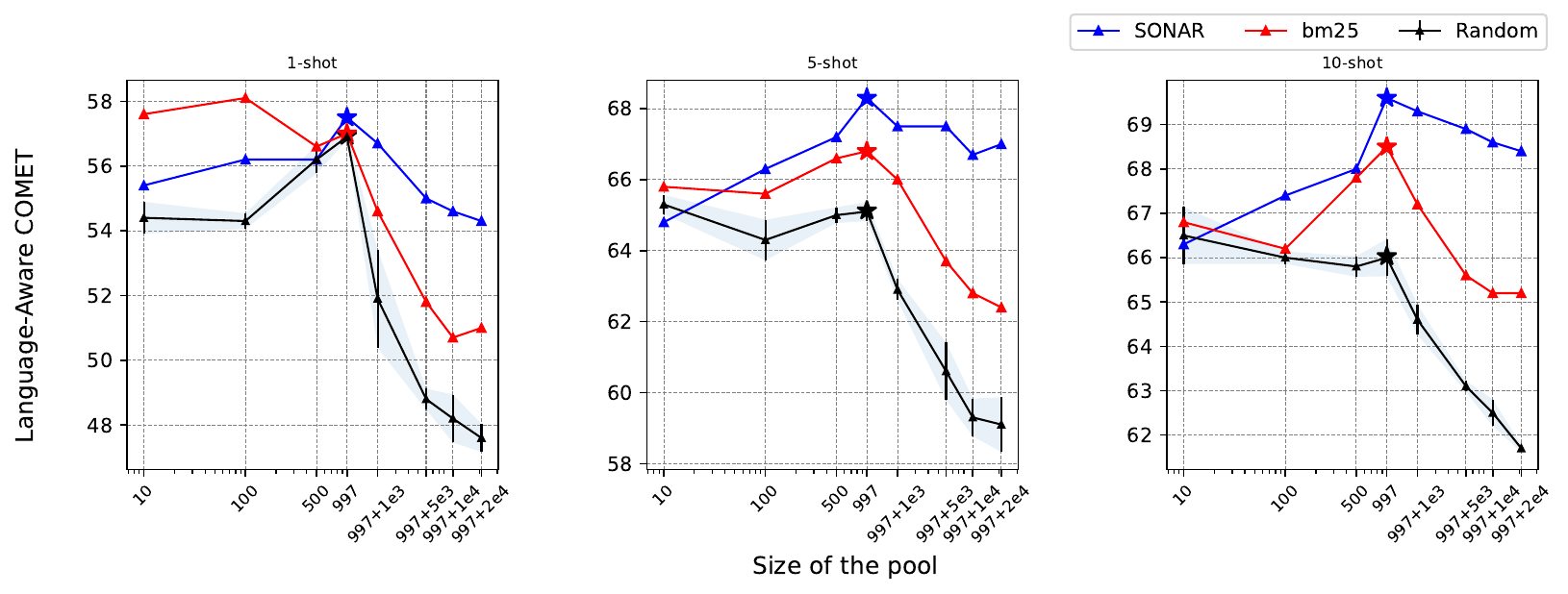}
     \caption{laCOMET scores for example retrieval with SONAR, BM25 and random sampling for various selection pool compositions for eng$\rightarrow$swh and BLOOM~7B1. The triangles correspond to the pool built either by shrinking $\mathcal{P}$ (taking the $N_1$ first pairs) or by extending it (with 
     the $N_2$ first pairs of $\mathcal{U}$). 
     The star indicates the initial pool, i.e.~the entire FLORES-200 dev set.}
     \label{fig:robustness-lacomet}
\end{figure*}

\begin{figure}[ht]
    \centering
    \includegraphics[width=0.45\textwidth]{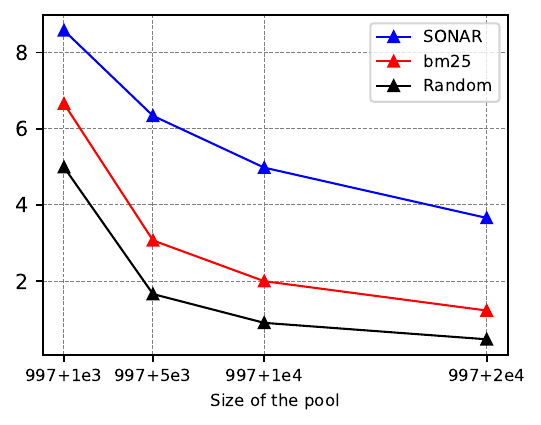}
    \caption{For each pool composition involving FLORES and NLLB samples, the average number of the 10 in-context examples belong to the FLORES-200 dev set when using SONAR, BM25, and random sampling.}
    \label{fig:choices}
\end{figure}

In order to gain more insights into which examples are being selected, we analyze, on average, what is the proportion of in-context examples belonging to the FLORES-200 dev set (i.e.~the highest quality examples) among the selected ones. We conduct the analysis in the 10-shot setting with BLOOM~7B1 and report the results in Figure~\ref{fig:choices}. We observe that despite having access to more samples, SONAR is more prone to selecting FLORES's samples than BM25. This suggests that SONAR is better at retrieving more high-quality samples even at the cost of sacrificing the $n$-gram-level similarity to the sentence of interest. This ability to query ``good sentences'' results in a greater resilience to noisy selection pools. Interestingly, as illustrated in Table~\ref{tab:average-similarity}, the average similarity scores between the retrieved examples in 10-shot increase with the size of the selection pool. This indicates that a larger pool improves the likelihood of retrieving relevant in-context demonstrations, although the quality of the retrieved examples is more important to generate good outputs.

\subsection{Scalability of example retrieval via similarity search} \label{scalability}

\begin{figure*}[t]
     \centering
     \includegraphics[width=0.95\textwidth]{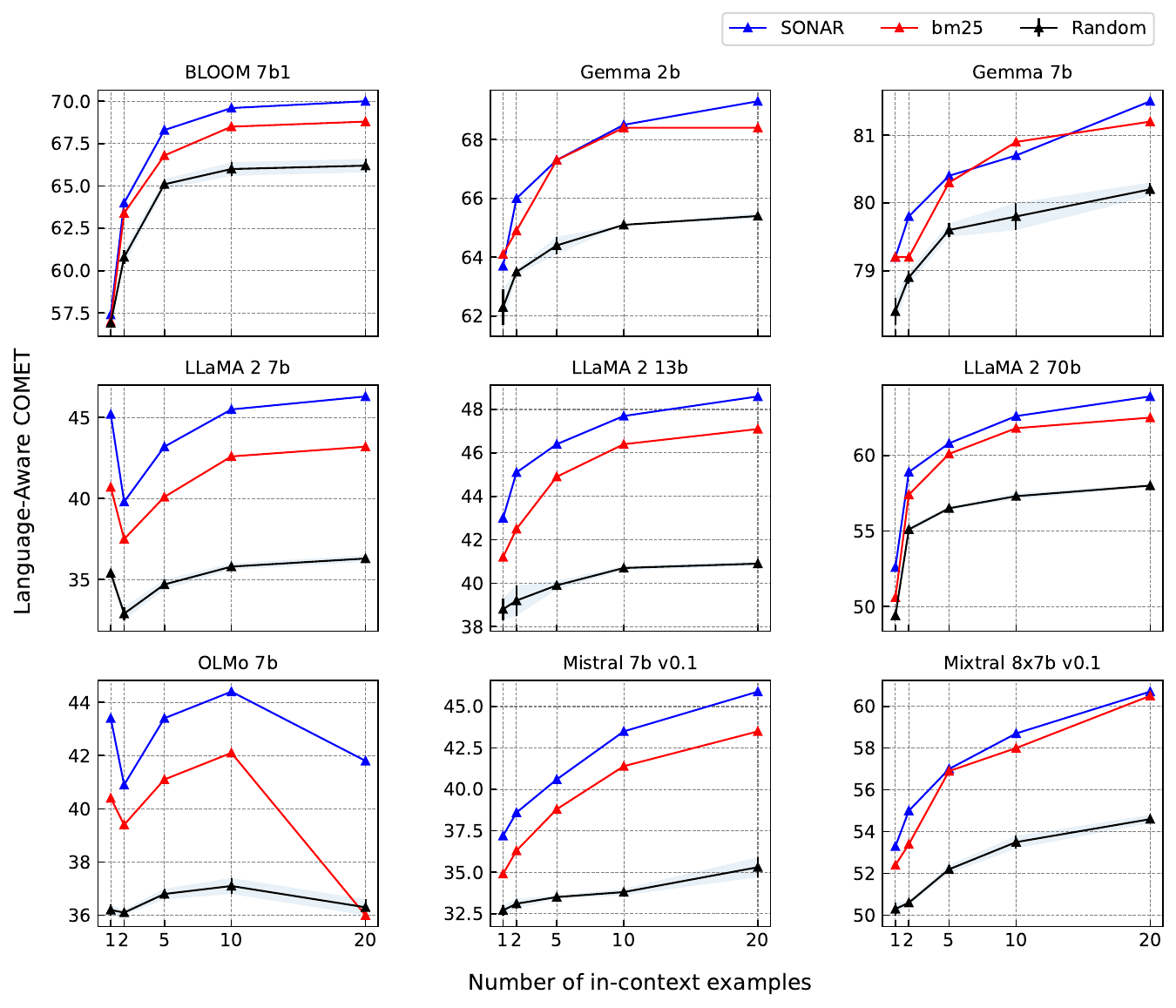}
     \caption{
     laCOMET scores of example retrieval with SONAR and BM25 compared to random sampling for the $k$-shot setting ($k \in \{1, 2, 5, 10, 20\}$) 
     for eng$\rightarrow$swh and nine LLMs. Note that for readability reasons, the Y-axis scales of the figures are not aligned.}
    \label{fig:compare}
\end{figure*}

We demonstrate that the advantages of example retrieval are observable across various scales by evaluating it on a range of LLMs with parameter counts ranging from 2B to 70B. Figure~\ref{fig:compare} highlights the efficacy of example retrieval when translating from English to Swahili. Most LLMs show a performance improvement of at least 4 laCOMET points between the use of SONAR and random sampling for example selection. Interestingly, we observe that even with 20 in-context examples, the gap with random sampling does not plummet; it continues to increase with the number of in-context examples.\footnote{OLMo~7B's performance drop in the 20-shot setting is caused by its short context length (2048) which makes most generations empty.} 
BM25 consistently outperforms random sampling but does not reach SONAR's laCOMET scores.

\section{Discussion}
\paragraph{Example selection via similarity search improves MT.}  
Our results for translation into French and German partially resonate with previous work by \citet{DBLP:conf/acl/VilarFCLRF23} and \citet{zhu2023multilingual}, as we reported a small range of improvement for these languages over random sampling for a high quality pool (between 0.1 and 0.5 laCOMET for most LLMs). However, our experiments on Swahili and Wolof show that example selection can yield significant gains for lower-resource languages. For these languages, and when the LLM's context length allowed it, we did not observe a plateau even at 20-shot as opposed to \citep{zhu2023multilingual}\footnote{We stopped at 20 because of the limited context length of some of our LLMs (e.g.~BLOOM~7B1, OLMo~7B), which would have resulted in truncated contexts and therefore have a negative impact on scores.}. In addition to a strong performance, example retrieval using SONAR is 
resilient with lower quality pools, outperforming the random baseline as well as the strong BM25 approach. This robustness is observed for both high- and low-resource directions in terms of BLEU and laCOMET.

\paragraph{What issues arise when prompting an LLM to translate into a low-resource language?} The zero-shot abilities of LLMs are sensitive to the template, as shown in Section~\ref{prompt-selection}. This is caused by two problems. First, there are instances where the model fails to understand the task and generates unrelated outputs (e.g.~multiple line breaks, a repetition of the end of the prompt in multiple languages or a continuation of the input sentence). Secondly, there is the inability to accurately perform the task, leading for example to the repetition of the input sentence (potentially with a few modifications), partial translation (e.g.~with repeating $n$-grams at the end) and translation in an incorrect language. Table~\ref{tab:issues} contains some examples of these issues produced by Mixtral~8x7B~v0.1. The first problem is generally minor when we have a good template, a high-resource language and a capable LLM (e.g.~template T5, French and Mistral~7B~v0.1 in Table~\ref{tab:prompt-result}). Moreover, it is mostly solved by using a 1-shot example. This is why there is a huge gap between 0-shot and 1-shot performance as pointed out by \citet{gpt-mt-2023}. Low-resource directions would require more shots, typically between 2 and 5. The second problem is more tenacious, particularly for low-resource directions. As the number of shot increases, the number of translations in the correct language increases and the number of empty translations decreases. However, the scores remain low.


\begin{table}[ht]
\centering\small
\begin{tabular}{p{2.5cm}p{4.5cm}}
\toprule
Source sentence & International sanctions have meant that new aircraft cannot be purchased.\\
0-shot translation (paraphrases source) & Senegal is under international sanctions, so new aircraft cannot be purchased. \\
\midrule
Source sentence & During his trip, Iwasaki ran into trouble on many occasions.\\
0-shot translation (wrong language) & Durant son voyage, Iwasaki a rencontré beaucoup de problèmes. \\
\bottomrule
\end{tabular}
\caption{Examples of 0-shot eng$\rightarrow$wol mistranslations by Mixtral~8x7B~v0.1\iffalse. In the first case, the model mostly outputs the same sentence as the source sentence, and hallucinates information (the underlined part). In the second case, the model translates the input sentence into French rather than Wolof\fi.}
\label{tab:issues}
\end{table}

\paragraph{Why does example selection via similarity search work?} The success of ICL depends on the ability of the LLM to understand the task and its ability to generate a qualitative output given an input. As explained earlier, the task understanding is mostly solved by using few-shot examples. Example selection via similarity search leads to gains in output quality by using qualitative demonstrations aimed at encouraging the LLM to generate higher quality outputs. The impact of example retrieval on the translation from English to French is noticeable at the phrasing level. It makes the LLMs employ different words compared to those used with random sampling to convey the same message. Additionally, it influences the translation of entities (e.g.~names of organizations, universities, stadiums, etc.), although we did not observe a consistent pattern in this regard. For translation into Wolof, we observed that example retrieval considerably impacts the rate at which the number of translations in the correct language increases,\footnote{See Appendix~\ref{appendix:distribution}.} partially explaining its superior performance. For translation into Swahili, example retrieval helps mitigate the uncontrollable generation of $n$-grams, and its impact on the phrasing is more pronounced than observed for French. The LLMs tend to generate more words in Swahili that are relevant to the context of the sentence to translate.

\section{Conclusions}
We have provided a systematic study of example selection via similarity search as a simple way to improve the MT capabilities of LLMs, comparing the translation quality of multiple open-source LLMs when using a range of different sentence embedding methods to select few-shot examples. We cover four translation directions covering high- and low-resource languages. Our results confirm previous results for high-resource languages that similarity search does not provide significant gains over random sampling. However, we show  that the strategy allows LLMs to demonstrate superior translation performance for mid- and low-resource languages. 
We validated these results across multiple scales of LLMs and example pool sizes. We also demonstrated that greater diversity in high-quality pools yields better results. Example retrieval is significantly more robust to quality heterogeneity, with sentence embeddings providing the highest resilience.



\section*{Limitations}
One inherent limitation of our work is the definition of the concept of similarity; it is a broad and polymorphous concept, and we choose to focus on semantics through the use of sentence embeddings (although it is likely that other aspects are also represented via sentence embeddings).  Although other approaches (e.g. more syntax-based) are also possible and would be interesting to explore in future work. 
Moreover, despite the gain observed when translating from English to Wolof, it is obvious that most LLMs struggle considerably with this language and other low-resource ones, and this should be a research direction to explore.


\section*{Acknowledgements}
This work was partly funded  by the last two authors' chairs in the PRAIRIE institute funded by the French national agency ANR as part of the ``Investissements d'avenir'' programme under the reference ANR-19-P3IA-0001.

\bibliography{anthology,custom}
\bibliographystyle{acl_natbib}

\appendix


\section{Implementation details}\label{appendix:experimental_details}

\subsection{Framework and hyperparameters}

All our experiments are done with beam search \citep{freitag-al-onaizan-2017-beam} and a beam size of~2. We use \texttt{vLLM} \citep{kwon2023efficient} for inference and generate with a maximum sentence length of 100 tokens. In zero-shot settings, we truncate the prediction at the first new line break and ignore any tokens generated afterwards.

\subsection{Models}

In Table~\ref{tab:url}, we list the links to the relevant resources used for experiments.
\begin{table*}[!ht]
    \centering\small
    \begin{tabular}{l l}
    \toprule
    \multicolumn{2}{c}{\textit{Datasets}} \\
    \midrule
    Flores-200 & \url{https://huggingface.co/datasets/facebook/flores} \\
    NLLB Full dataset & \url{https://huggingface.co/datasets/allenai/nllb}  \\
    \midrule
    \multicolumn{2}{c}{\textit{Models evaluated}} \\
    \midrule
    BLOOM~7B1 & \url{https://huggingface.co/bigscience/bloom-7b1} \\
    OLMo~7B & \url{https://huggingface.co/allenai/OLMo-7B}\\
    Gemma~2B & \url{https://huggingface.co/google/gemma-2b} \\
    Gemma~7B & \url{https://huggingface.co/google/gemma-7b} \\
    LLaMA~2~7B & \url{https://huggingface.co/meta-llama/Llama-2-7b-hf} \\
    LLaMA~2~13B & \url{https://huggingface.co/meta-llama/Llama-2-13b-hf} \\
    LLaMA~2~70B & \url{https://huggingface.co/TheBloke/Llama-2-70B-AWQ} \\
    Mistral~7B~v0.1 & \url{https://huggingface.co/mistralai/Mistral-7B-v0.1}\\
    Mixtral~8x7B~v0.1 & \url{https://huggingface.co/TheBloke/mixtral-8x7B-v0.1-AWQ} \\
    RoBERTa &  \url{https://huggingface.co/FacebookAI/roberta-large}\\
    \midrule
    \multicolumn{2}{c}{\textit{Sentence embeddings}} \\
    \midrule
    Cohere & \texttt{embed-multilingual-v3.0}\\
    E5 & \url{https://huggingface.co/intfloat/multilingual-e5-large} \\
    LaBSE & \url{https://huggingface.co/sentence-transformers/LaBSE} \\
    Laser 2 & \url{https://github.com/facebookresearch/LASER} \\ 
    SONAR & \url{https://github.com/facebookresearch/SONAR} \\
    \bottomrule
    \end{tabular}
    \caption{Links to datasets, benchmarks and models.}
    \label{tab:url}
\end{table*}

\section{Additional results}\label{app:additional_results}

\subsection{Impact of in-context example order} \label{appendix:reverse}

We investigated how the ranking of in-context examples impacts translation performance. Given the huge number of permutations possible, we could not evaluate each of them. Instead, we compared the current order to its direct opposite (i.e.~ranking the retrieved in-context examples from the least to the most similar starting from the source sentence). The results, given in Table~\ref{tab:reverse} show that there is no significant difference in performance between the two orders.

\begin{table*}[ht]
\centering\small
\begin{tabular}{llcccccccccccc}
\toprule
& \multicolumn{1}{c}{\bf }  & \multicolumn{3}{c}{\bf eng$\rightarrow$fra} & \multicolumn{3}{c}{\bf eng$\rightarrow$deu} & \multicolumn{3}{c}{\bf eng$\rightarrow$swh} & \multicolumn{3}{c}{\bf eng$\rightarrow$wol}\\
 & & 1 & 5 & 10 & 1 & 5 & 10 & 1 & 5 & 10 & 1 & 5 & 10\\
\midrule
\multirow{2}{*}{BLEU} & Original & \bf 42.9 & \bf 47.5 & \bf 48.0 & \bf 12.6 & \bf 14.9 & \bf 15.2 & \bf 8.6 & \bf 12.0 & 12.7 & \bf 2.2 & \bf 2.9 & \bf 3.0 \\
& Reverse & \bf 42.9 & 47.0 & 47.8 & \bf 12.6 & \bf 14.9 & \bf 15.2 & \bf 8.6 & 11.7 & \bf 13.1 & \bf 2.2 & 2.8 & \bf 3.0 \\
\midrule
\multirow{2}{*}{COMET} & Original & \bf 84.9 & \bf 86.8 & \bf 86.6 & \bf 58.9 & \bf 60.7 & \bf 61.3 & \bf 64.5 & \bf 69.5 & 70.4 & \bf 52.0 & 51.6 & \bf 52.5 \\
& Reverse & \bf 84.9 & 86.6 & \bf 86.6 & \bf 58.9 & 60.2 & \bf 61.3 & \bf 64.5 & 69.2 & \bf 70.5 & \bf 52.0 & \bf 51.7 & \bf 52.5 \\
\midrule
\multirow{2}{*}{laCOMET} & Original & \bf 79.8 & \bf 86.8 & \bf 86.6 & \bf 55.3 & \bf 60.1 & 60.8 & \bf 57.4 & \bf 68.3 & 69.6 & \bf 50.2 & \bf 50.4 & \bf 51.6 \\
& Reverse & \bf 79.8 & 86.6 & \bf 86.6 & \bf 55.3 & 59.6 & \bf 60.9 & \bf 57.4 & 67.8 & \bf 69.7 & \bf 50.2 & 50.3 & \bf 51.6 \\
\bottomrule
\end{tabular}
\caption{Impact of the ordering of in-context examples (Original: most to least similar, Reverse: least to most similar) in $k$-shot settings ($k \in \{1, 5, 10\}$) on translation quality (BLEU, COMET and laCOMET) with BLOOM~7B1 as the translator and SONAR as the example retriever.}
\label{tab:reverse}
\end{table*}

\subsection{Overlap between sentence embeddings} \label{appendix:overlap}
\begin{figure}[ht]
     \centering
     \includegraphics[width=\linewidth]{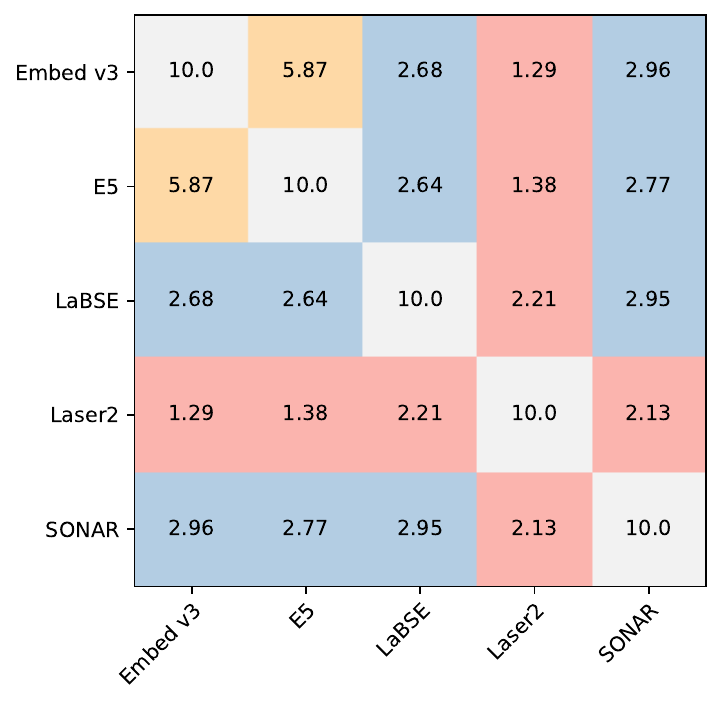}
    \caption{Average number of retrieved examples in common  between sentence embedding methods (10-shot).}
    \label{fig:intersection}
\end{figure}

Motivated by the low variability in performance observed between the sentence embeddings in Table~\ref{tab:lacomet-comparison},  we analyzed the degree of overlap in the choices made by the different sentence embedding methods by calculating the average intersection between the top 10 pairs retrieved (in $\mathcal{P}$) between methods (the pool being the Flores-200 devtest set). The results in Figure~\ref{fig:intersection} show that each method retrieved a distinct set of examples, with most overlap seen between E5 and Embed~v3 with an average of 5.87 examples in common per top 10.

\subsection{BLEU and COMET results} \label{appendix:bleu-comet}

As mentioned previously we additionally present results with BLEU (Table~\ref{tab:bleu-se} and Table~\ref{tab:bleu-comparison}) and COMET (Table~\ref{tab:comet-se} and Table~\ref{tab:comet-comparison}) for transparency reasons. The results show the same pattern as the laCOMET results shown in the main part of the paper. Example retrieval with sentence embeddings outperforms random sampling in all scenarios.

\begin{table*}[ht]
\centering\small
\begin{center}
\begin{tabular}{lcccccccccccc}
\toprule
\multicolumn{1}{c}{\bf }  & \multicolumn{3}{c}{\bf eng $\rightarrow$ fra} & \multicolumn{3}{c}{\bf eng $\rightarrow$ deu} & \multicolumn{3}{c}{\bf eng $\rightarrow$ swh} & \multicolumn{3}{c}{\bf eng $\rightarrow$ wol}\\
 {} & 1 & 5 & 10 & 1 & 5 & 10 & 1 & 5 & 10 & 1 & 5 & 10\\
\midrule
BLOOM~7B1 \\
\midrule
Embed v3 & 42.3 & 47.0 & 47.5 & 12.4 & 14.7 & 15.1 & \bf 8.9 & \textbf{12.3} & \textbf{12.7} & 1.8 & 2.3 & 2.5 \\
E5 & 42.7 & 47.2 & 47.9 & 12.5 & \bf 14.9 & \bf 15.3 & 8.6 & 12.1 & 12.5 & 1.9 & 2.4 & 2.6 \\
LaBSE & 42.5 & 47.3 & 47.8 & 12.6 & \bf 14.9 & 15.2 & 8.7 & 11.7 & 12.3 & \bf 2.4 & 2.6 & 2.9 \\
LASER2 & 42.1 & 47.4 & 47.9 & \textbf{12.8} & 14.6 & 15.0 & 8.6 & 11.6 & 11.7 & \textbf{2.4} & 2.8 & 2.9 \\
SONAR & \bf 42.9 & \bf 47.5 & \bf 48.0 & 12.6 & \bf 14.9 & 15.2 & 8.6 & 12.0 & \bf 12.7 & 2.2 & \textbf{2.9} & \textbf{3.0} \\
Random & 40.8 & 46.7 & 47.2 & 12.3 & 13.9 & 14.0 & 8.2 & 10.5 & 11.0 & 0.9 & 1.6 & 1.9\\
\midrule
Mistral~7B~v0.1\\
\midrule
Embed v3 & 47.3 & 48.4 & 48.8 & 36.4 & 38.0 & \textbf{38.6} & \textbf{3.6} & \textbf{4.9} & 5.4 & 2.8 & 3.3 & 3.7 \\
E5 & 46.9 & 48.5 & 48.7 & 36.4 & 37.9 & 38.2 & 3.5 & 4.7 & \textbf{5.5} & 2.8 & 3.3 & 3.3 \\
LaBSE & 47.4 & 48.8 & 49.0 & 36.5 & 37.8 & 37.9 & 3.3 & 4.6 & 5.1 & 3.2 & 3.3 & \textbf{3.8} \\
LASER2 & \textbf{47.5} & 48.8 & 49.0 & 36.3 & 37.4 & 37.7 & 3.1 & 4.1 & 4.7 & \textbf{3.3} & 3.3 & 3.6 \\
SONAR & 47.4 & \textbf{49.0} & \textbf{49.2} & \textbf{36.6} & \textbf{38.1} & 38.2 & 3.5 & 4.6 & 5.4 & 3.2 & \textbf{3.4} & 3.7 \\
Random & 47.2 & 48.0 & 48.4 & 36.1 & 37.3 & 37.5 & 2.8 & 2.8 & 2.9 & 2.4 & 2.3 & 2.7 \\
\midrule
LLaMA~2~7B \\
\midrule
Embed v3 & 44.5 & 45.8 & 46.1 & 34.7 & 35.3 & 35.4 & 2.9 & 4.1 & 4.4 & 2.0 & 3.2 & 3.4 \\
E5 & 44.8 & \textbf{46.0} & \textbf{46.3} & \bf 34.8 & \textbf{35.9} & \textbf{35.7} & 3.1 & 3.8 & 4.4 & 2.0 & 3.0 & 3.1 \\
LaBSE & 44.4 & 45.3 & 46.0 & \bf 34.8 & 35.6 & 35.4 & \bf 3.2 & \bf 4.2 & 4.3 & 2.5 & \bf 3.6 & \bf 3.7 \\
LASER2 & 44.8 & 45.6 & 46.1 & 34.6 & 35.7 & \bf 35.7 &  3.1 & 3.6 & 4.0 & \textbf{2.6} & \textbf{3.6} & 3.6 \\
SONAR & \textbf{44.9} & 45.5 & 46.0 & 34.5 & 35.7 & \bf 35.7 & 3.1 & \textbf{4.2} & \textbf{4.6} & 2.1 & 3.4 & 3.7 \\
Random & 44.6 & 45.2 & 45.4 & 34.1 & 35.2 & 35.5 & 2.4 & 2.8 & 2.8 & 1.3 & 2.1 & 2.3 \\
\midrule
Gemma~7B \\
\midrule
Embed v3 & 52.0 & 52.7 & 53.4 & 42.0 & 42.5 & 42.8 & 26.4 & \textbf{28.5} & \textbf{29.4} & 1.9 & 3.0 & 3.5 \\
E5 & 51.8 & 52.6 & 53.3 & 41.8 & 42.7 & 42.9 & 26.6 & 28.2 & 29.1 & 2.1 & 3.1 & 3.6 \\
LaBSE & \textbf{52.2} & 53.1 & 53.2 & \textbf{42.4} & 42.7 & 42.8 & \textbf{26.8} & 28.4 & 29.2 & 2.1 & 3.1 & \bf 3.7 \\
LASER2 & 52.0 & \textbf{53.2} & 53.4 & 41.7 & 42.3 & 42.3 & 26.6 & 28.0 & 28.5 & 1.9 & 3.1 & 3.6 \\
SONAR & \bf 52.2 & 53.1 & \textbf{53.5} & 41.8 & \textbf{42.8} & \textbf{43.3} & 26.5 & 28.1 & 28.6 & \textbf{2.2} & \textbf{3.2} & \textbf{3.7} \\
Random & 52.0 & 52.8 & 53.0 & 41.8 & 42.4 & 42.5 & 25.8 & 26.7 & 27.0 & 1.4 & 2.0 & 2.4 \\
\bottomrule
\end{tabular}
\end{center}
\caption{BLEU scores for $k$-shot ($k \in \{1, 5, 10\}$) example retrieval with different sentence embeddings.}
\label{tab:bleu-se}
\end{table*}

\begin{table*}[ht]
\centering\small
\begin{tabular}{lcccccccccccc}
\toprule
\multicolumn{1}{c}{\bf }  & \multicolumn{3}{c}{\bf eng $\rightarrow$ fra} & \multicolumn{3}{c}{\bf eng $\rightarrow$ deu} & \multicolumn{3}{c}{\bf eng $\rightarrow$ swh} & \multicolumn{3}{c}{\bf eng $\rightarrow$ wol}\\
 {} & 1 & 5 & 10 & 1 & 5 & 10 & 1 & 5 & 10 & 1 & 5 & 10\\
\midrule
BLOOM~7B1 \\
\midrule
Embed v3 & 84.6 & 86.7 & \bf 86.7 & \textbf{59.0} & 60.6 & \bf 61.3 & \textbf{65.1} & \textbf{69.7} & 70.2 & \textbf{52.4} & 51.4 & 52.0\\
E5 & \textbf{85.0} & 86.6 & \bf 86.7 & 58.7 & 60.5 & 61.0 & \bf 65.1 & 69.6 & 70.2 & 52.7 & \bf 52.3 & 51.9 \\
LaBSE & 84.7 & 86.7 & \bf 86.7 & 58.8 & 60.4 & 61.2 & 64.5 & 69.2 & 69.9 & 52.0 & \bf 52.3 & \textbf{53.1}\\
LASER2 & 84.8 & 86.6 & \bf 86.7 & 58.8 & 60.3 & 60.3 & 64.1 & 68.9 & 68.9 & 51.5 & 51.4 & 52.3 \\
SONAR & 84.9 & \textbf{86.8} & 86.6 & 58.9 & \textbf{60.7} & \textbf{61.3} & 64.5 & 69.5 & \textbf{70.4} & 52.0 & 51.6 & 52.5 \\
Random & 84.3 & 86.5 & 86.6 & 58.0 & 58.5 & 58.7 & 64.0 & 67.7 & 67.9 & 49.0 & 47.3 & 48.3 \\
\midrule
Mistral~7B~v0.1\\
\midrule
Embed v3 & \bf 86.6 & \bf 87.0 & 87.0 & 84.8 & 85.8 & 86.0 & \textbf{41.8} & 43.0 & \textbf{45.1} & 45.2 & 48.2 & 48.6 \\
E5 & 86.4 & \bf 87.0 & 86.9 & 84.9 & 85.7 & 85.8 & 41.6 & \textbf{43.3} & 44.9 & \textbf{45.5} & \textbf{48.5} & 48.5 \\
LaBSE & 86.5 & 86.9 & 87.0 & 84.9 & 85.7 & 85.9 & 41.3 & 42.2 & 43.7 & 45.5 & 47.1 & 48.8 \\
LASER2 & 86.5 & \bf 87.0 & 87.0 & \bf 85.0 & 85.8 & 85.8 & 39.7 & 40.1 & 41.9 & 43.3 & 47.0 & 47.6 \\
SONAR & 86.3 & \textbf{87.0} & \textbf{87.1} & \textbf{85.0} & \textbf{85.9} & \textbf{86.1} & 41.4 & 42.8 & 45.1 & 45.3 & 48.4 & \textbf{49.0} \\
Random & 86.4 & 86.7 & 86.7 & 84.7 & 85.7 & 85.7 & 38.1 & 36.6 & 36.7 & 39.3 & 40.3 & 42.4 \\
\midrule
LLaMA~2~7B \\
\midrule
Embed v3 & 85.8 & 86.1 & 86.3 & 84.2 & 85.0 & 85.0 & \textbf{48.7} & \textbf{45.8} & \textbf{46.5} & 48.5 & 50.0 & \textbf{50.4} \\
E5 & 85.8 & \textbf{86.2} & \textbf{86.4} & \textbf{84.4} & 85.2 & 85.2 & 48.3 & 45.1 & 46.5 & \textbf{48.9} & \textbf{50.2} & 49.7 \\
LaBSE & 85.8 & 86.0 & 86.2 & 84.4 & 85.2 & 85.1 & 47.6 & 44.9 & 45.9 & 48.2 & 49.0 & 49.6 \\
LASER2 & 85.8 & \bf 86.2 & 86.2 & 84.1 & 85.1 & 85.3 & 44.8 & 42.3 & 43.3 & 47.2 & 48.4 & 48.2 \\
SONAR & \textbf{85.9} & 86.1 & 86.3 & 84.2 & \textbf{85.3} & \textbf{85.4} & 48.5 & 44.8 & 46.4 & 47.9 & \bf 50.2 & 50.3 \\
Random & 85.6 & 85.9 & 86.0 & 84.1 & 84.9 & 85.1 & 40.2 & 37.5 & 37.9 & 44.2 & 42.2 & 43.2 \\
\midrule
Gemma~7B \\
\midrule
Embed v3 & 87.6 & \bf 88.0 & \bf 88.1 & 86.9 & 87.3 & 87.5 & 79.4 & 80.8 & \textbf{81.4} & 42.2 & 46.6 & 49.0 \\
E5 & 87.5 & 87.9 & \bf 88.1 & 87.0 & 87.4 & 87.6 & \textbf{79.7} & 80.6 & 81.2 & \textbf{42.8} & 46.5 & 49.4 \\
LaBSE & \textbf{87.8} & 87.9 & 88.0 & \bf 87.1 & \bf 87.6 & 87.4 & 79.4 & \textbf{80.8} & 81.2 & 41.0 & 46.2 & 49.1 \\
LASER2 & 87.6 & 87.9 & 87.9 & \bf 87.1 & 87.4 & 87.2 & 79.6 & 80.6 & 80.6 & 40.3 & 45.8 & 48.7 \\
SONAR & 87.5 & \textbf{88.0} & \textbf{88.1} & 86.9 & \textbf{87.6} & \textbf{87.6} & 79.5 & 80.5 & 80.7 & 42.1 & \textbf{46.9} & \textbf{49.6} \\
Random & 87.6 & 87.9 & 88.0 & 86.8 & 87.2 & 87.3 & 78.7 & 79.8 & 79.9 & 36.2 & 39.9 & 42.6 \\
\bottomrule
\end{tabular}
\caption{COMET scores for $k$-shot ($k \in \{1, 5, 10\}$) example retrieval with different sentence embeddings.}
\label{tab:comet-se}
\end{table*}

\begin{table*}[ht]
\centering\small
\begin{center}
\begin{tabular}{lcccccccccccc}
\toprule
\multicolumn{1}{c}{\bf }  & \multicolumn{3}{c}{\bf eng$\rightarrow$fra} & \multicolumn{3}{c}{\bf eng$\rightarrow$deu} & \multicolumn{3}{c}{\bf eng$\rightarrow$swh} & \multicolumn{3}{c}{\bf eng$\rightarrow$wol}\\
 {} & 1 & 5 & 10 & 1 & 5 & 10 & 1 & 5 & 10 & 1 & 5 & 10\\
\midrule
BLOOM~7B1 \\
\midrule
SONAR & 42.9 & 47.5 & 48.0 & \bf 12.6 & 14.9 & 15.2 & 8.6 & \textbf{12.0} & 12.7 & \textbf{2.2} & \textbf{2.9} & \textbf{3.0} \\
BM25 & 41.1 & \textbf{47.7} & \textbf{48.1} & \bf 12.6 & \bf 15.1 & 15.2 & 8.8 & 11.6 & \textbf{12.9} & 1.8 & 2.3 & 2.8 \\
R-BM25 & \bf 43.3 & 46.1 & 46.8 & 12.4 & 13.7 & 13.8 & 7.9 & 10.2 & 10.7 & 1.2 & 1.5 & 2.1 \\
BLEU & 41.5 & 47.4 & 47.6 & 12.4 & 14.8 & \bf 15.3 & \bf 8.9 & 11.4 & 12.2 & 1.5 & 2.5 & 2.8 \\
RoBERTa & 41.3 & 46.6 & 47.6 & 12.4 & 14.2 & 14.0 & 8.6 & 10.5 & 11.4 & 1.6 & 2.2 & 2.2 \\
Random & 40.8 & 46.7 & 47.2 & 12.3 & 13.9 & 14.0 & 8.2 & 10.5 & 11.0 & 0.9 & 1.6 & 1.9\\
\midrule
Mistral~7B~v0.1\\
\midrule
SONAR & 47.4 & \textbf{49.0} & \textbf{49.2} & 36.6 & \textbf{38.1} & \textbf{38.2} & \bf 3.5 & 4.6 & 5.4 & \textbf{3.2} & \textbf{3.4} & \textbf{3.7} \\
BM25 & 47.7 & 48.6 & 49.0 & 36.5 & 37.9 & 38.1 & 3.4 & \textbf{5.0} & \textbf{5.7} & 2.8 & 3.3 & 3.4 \\
R-BM25 & 47.5 & 47.8 & 48.3 & 36.4 & 36.9 & 36.9 & 2.6 & 2.9 & 2.9 & 2.5 & 2.6 & 2.9 \\
BLEU & \textbf{47.9} & 48.5 & 49.0 & \textbf{36.8} & 37.6 & 37.8 & \textbf{3.5} & 4.5 & 4.8 & 2.6 & 2.9 & 3.2 \\
RoBERTa & 47.6 & 48.6 & 49.0 & 36.3 & 37.5 & 37.8 & 2.9 & 3.3 & 3.8 & 2.6 & 2.7 & 2.8 \\
Random & 47.2 & 48.0 & 48.4 & 36.1 & 37.3 & 37.5 & 2.8 & 2.8 & 2.9 & 2.4 & 2.3 & 2.7 \\
\midrule
LLaMA~2~7B \\
\midrule
SONAR & 44.9 & 45.5 & 46.0 & 34.5 & 35.7 & 35.7 & \textbf{3.1} & \textbf{4.2} & 4.6 & \textbf{2.1} & \textbf{3.4} & \textbf{3.7} \\
BM25 & \textbf{45.0} & 45.9 & 46.1 & 34.4 & \textbf{35.8} & \textbf{36.1} & \bf 3.1 & 4.0 & \textbf{4.7} & 1.8 & 3.0 & 3.0 \\
R-BM25 & 44.5 & 45.2 & 45.0 & 33.8 & 34.9 & 35.1 & 2.5 & 2.8 & 2.9 & 1.2 & 2.3 & 2.4 \\
BLEU & 44.8 & \textbf{46.0} & \textbf{46.4} & \textbf{34.6} & 35.6 & 35.7 & 3.0 & 3.9 & 4.3 & 1.7 & 2.7 & 3.1 \\
RoBERTa & 44.7 & 45.8 & 45.9 & \bf 34.6 & 35.6 & 35.9 & 2.7 & 3.1 & 3.5 & 1.4 & 2.5 & 2.6 \\
Random & 44.6 & 45.2 & 45.4 & 34.1 & 35.2 & 35.5 & 2.4 & 2.8 & 2.8 & 1.3 & 2.1 & 2.3 \\
\midrule
Gemma~7B \\
\midrule
SONAR & 52.2 & 53.1 & 53.5 & 41.8 & 42.8 & \textbf{43.3} & 26.5 & 28.1 & 28.6 & \textbf{2.2} & \textbf{3.2} & \textbf{3.7} \\
BM25 & 52.3 & 52.9 & 52.5 & 41.6 & 42.7 & 42.6 & \textbf{26.8} & \textbf{28.4} & \textbf{29.2} & 1.8 & 2.9 & 3.5 \\
R-BM25 & 52.6 & 52.8 & 52.7 & 41.4 & 41.7 & 41.6 & 25.6 & 26.8 & 27.0 & 1.4 & 2.1 & 2.4 \\
BLEU & \textbf{52.7} & \textbf{53.3} & 53.2 & \textbf{42.3} & 42.6 & 42.9 & 26.6 & 28.1 & 28.6 & 1.8 & 2.7 & 3.2 \\
RoBERTa & 51.9 & 53.2 & \bf 53.6 & 41.9 & \textbf{42.9} & 42.9 & 26.1 & 27.4 & 27.3 & 1.7 & 2.4 & 2.8\\
Random & 52.0 & 52.8 & 53.0 & 41.8 & 42.4 & 42.5 & 25.8 & 26.7 & 27.0 & 1.4 & 2.0 & 2.4 \\
\bottomrule
\end{tabular}
\end{center}
\caption{Comparison of $k$-shot ($k \in \{1, 5, 10\}$) example retrieval with SONAR to baseline methods (BLEU). 
}
\label{tab:bleu-comparison}
\end{table*}

\begin{table*}[ht]
\centering\small
\begin{center}
\begin{tabular}{lcccccccccccc}
\toprule
\multicolumn{1}{c}{\bf }  & \multicolumn{3}{c}{\bf eng $\rightarrow$ fra} & \multicolumn{3}{c}{\bf eng $\rightarrow$ deu} & \multicolumn{3}{c}{\bf eng $\rightarrow$ swh} & \multicolumn{3}{c}{\bf eng $\rightarrow$ wol}\\
 {} & 1 & 5 & 10 & 1 & 5 & 10 & 1 & 5 & 10 & 1 & 5 & 10\\
\midrule
BLOOM~7B1 \\
\midrule
SONAR & 84.9 & \bf 86.8 & 86.6 & \bf 58.9 & \bf 60.7 & \bf 61.3 & 64.5 & \bf 69.5 &\bf 70.4 & \bf 52.0 & \bf 51.6 & \bf 52.5 \\
BM25 & 84.6 & 86.6 & 86.7 & 58.3 & 60.1 & 60.1 & 64.6 & 68.4 & 69.5 & 51.3 & 50.8 & 51.6 \\
R-BM25 & \bf 85.2 & 86.4 & 86.5 & 58.0 & 58.3 & 59.2 & 63.2 & 67.4 & 67.8 & 46.7 & 46.4 & 47.7 \\
BLEU & 84.4 & 86.7 & 86.6 & 58.1 & 59.9 & 60.4 & 64.4 & 68.1 & 68.8 & 51.4 & 50.8 & 51.9 \\
RoBERTa & 84.5 & 86.7 & \bf 86.8 & 58.5 & 59.8 & 59.1 & \bf 64.8 & 67.7 & 68.5 & 51.7 & 50.7 & 50.8 \\
Random & 84.3 & 86.5 & 86.6 & 58.0 & 58.5 & 58.7 & 64.0 & 67.7 & 67.9 & 49.0 & 47.3 & 48.3 \\
\midrule
Mistral~7B~v0.1\\
\midrule
SONAR & 86.3 & \bf 87.0 & \bf 87.1 & \bf 85.0 & \bf 85.9 & \bf 86.1 & \bf 41.4 & \bf 42.8 & \bf 45.1 & \bf 45.3 & \bf 48.4 & \bf 49.0 \\
BM25 & \bf 86.6 & 86.8 & 86.9 & 84.8 & 85.7 & 85.9 & 40.1 & 41.1 & 43.2 & 43.6 & 45.8 & 47.6\\
R-BM25 & 86.5 & 86.7 & 86.7 & 84.9 & 85.6 & 85.8 & 37.6 & 36.5 & 36.6 & 38.3 & 39.1 & 41.2 \\
BLEU & \bf 86.6 & 86.9 & 86.9 & 84.9 & 85.7 & 85.9 & 39.8 & 39.9 & 41.1 & 42.8 & 44.8 & 46.7\\
RoBERTa & 86.5 & 86.9 & 87.0 & 84.9 & 85.6 & 86.0 & 39.1 & 38.2 & 39.3 & 42.5 & 44.5 & 45.7 \\
Random & 86.4 & 86.7 & 86.7 & 84.7 & 85.7 & 85.7 & 38.1 & 36.6 & 36.7 & 39.3 & 40.3 & 42.4 \\
\midrule
LLaMA~2~7B \\
\midrule
SONAR & \bf 85.9 & 86.1 & \bf 86.3 & 84.2 & \bf 85.3 & \bf 85.4 & \bf 48.5 & \bf 44.8 & \bf 46.4 & \bf 47.9 & \bf 50.2 & \bf 50.3\\
BM25 & 85.7 & 86.1 & 86.2 & 84.0 & 85.0 & 85.1 & 44.4 & 42.3 & 43.9 & 46.8 & 48.0 & 48.6\\
R-BM25 & 85.6 & 86.0 & 85.8 & 84.0 & 85.1 & 85.0 & 39.2 & 37.2 & 37.3 & 40.7 & 38.7 & 40.9 \\
BLEU & 85.6 & 86.0 & 86.1 & \bf 84.3 & 85.0 & 85.0 & 43.2 & 41.1 & 41.8 & 46.3 & 46.9 & 47.7\\
RoBERTa & 85.7 & \bf 86.2 & 86.0 & \bf 84.3 & 85.1 & 85.3 & 44.2 & 40.1 & 41.3 & 47.0 & 47.0 & 47.2 \\
Random & 85.6 & 85.9 & 86.0 & 84.1 & 84.9 & 85.1 & 40.2 & 37.5 & 37.9 & 44.2 & 42.2 & 43.2 \\
\midrule
Gemma~7B \\
\midrule
SONAR & 87.5 & \bf 88.0 & \bf 88.1 & 86.9 & \bf 87.6 & \bf 87.6 & \bf 79.5 & \bf 80.5 & 80.7 & \bf 42.1 & \bf 46.9 & \bf 49.6 \\
BM25 & 87.6 & \bf 88.0 & 87.7 & 86.9 & 87.3 & 87.0 & 79.4 & \bf 80.5 & \bf 80.9 & 39.8 & 45.4 & 48.5 \\
R-BM25 & 87.6 & 87.9 & 87.7 & 86.9 & 87.1 & 86.8 & 78.7 & 79.9 & 79.8 & 34.6 & 38.3 & 40.7 \\
BLEU & \bf 87.7 & 87.9 & \bf 88.1 & \bf 87.1 & 87.5 & 87.4 & 79.2 & \bf 80.5 & 80.3 & 39.5 & 44.2 & 47.0 \\
RoBERTa & 87.5 & 88.1 & \bf 88.1 & 86.9 & 87.4 & 87.4 & 79.0 & 80.2 & 80.1 & 39.8 & 42.5 & 45.3 \\
Random & 87.6 & 87.9 & 88.0 & 86.8 & 87.2 & 87.3 & 78.7 & 79.8 & 79.9 & 36.2 & 39.9 & 42.6 \\
\bottomrule
\end{tabular}
\end{center}
\caption{Comparison of $k$-shot ($k \in \{1, 5, 10\}$) example retrieval with SONAR to baseline methods (COMET). 
}
\label{tab:comet-comparison}
\end{table*}

\subsection{Additional results for other LLMs}\label{appendix:additional-results}

In Tables~\ref{tab:more-lacomet-se} and~\ref{tab:more-lacomet-comparison}, we provide the laCOMET scores for five additional LLMs: Gemma~2B, OLMo~7B, LLaMA~2~13B, LLaMA~2~70B, and Mixtral~8x7B~v0.1. We observe the same results as with BLOOM~7B1, Mistral~7B~v0.1, LLaMA~2~7B and Gemma~7B. Example retrieval with sentence embeddings outperforms random sampling at all scales, with the delta being higher when translating into Swahili and Wolof. SONAR is overall the best alternative, followed by BM25.

\begin{table*}[ht]
\centering\small
\begin{center}
\begin{tabular}{lcccccccccccc}
\toprule
\multicolumn{1}{c}{\bf }  & \multicolumn{3}{c}{\bf eng$\rightarrow$fra} & \multicolumn{3}{c}{\bf eng$\rightarrow$deu} & \multicolumn{3}{c}{\bf eng$\rightarrow$swh} & \multicolumn{3}{c}{\bf eng$\rightarrow$wol}\\
 {} & 1 & 5 & 10 & 1 & 5 & 10 & 1 & 5 & 10 & 1 & 5 & 10\\
\midrule
Gemma~2B \\
\midrule
Embed v3 & 84.7 & \bf 85.3 & \bf 85.4 & 82.0 & 83.1 & 83.3 & 63.9 & \bf 68.0 & \bf 68.6 & \bf 39.1 & 45.7 & 47.1 \\
E5 & 84.6 & 85.1 & \bf 85.4 & 82.2 & 83.2 & 83.2 & \bf 64.1 & 67.8 & 68.1 & 38.6 & \bf 45.8 & 47.2 \\
LaBSE & \bf 84.8 & 85.2 & \bf 85.4 & \bf 82.2 & \bf 83.4 & 83.4 & 64.0 & 67.0 & 68.1 & 36.3 & 44.7 & 46.9 \\
LASER2 & 84.6 & 85.0 & 85.0 & 82.0 & 83.1 & 83.2 & 63.7 & 66.3 & 67.4 & 32.5 & 42.7 & 44.9 \\
SONAR & \bf 84.8 & 85.2 & 85.3 & 82.0 & 83.2 & \bf 83.5 & 63.7 & 67.3 & 68.5 & 38.2 & 44.5 & \bf 47.4 \\
Random & 84.6 & 84.7 & 84.9 & 81.7 & 82.7 & 83.0 & 62.3 & 64.4 & 65.1 & 26.8 & 35.2 & 37.7 \\
\midrule
OLMo~7B \\
\midrule
Embed v3 & \bf 81.0 & 81.1 & 81.2 & \bf 75.0 & 75.7 & 75.6 & 43.2 & 43.0 & 44.2 & 40.4 & \bf 42.1 & 43.6 \\
E5 & \bf 81.0 & \bf 81.4 & 81.3 & 74.7 & 75.9 & 76.0 & 42.9 & 42.0 & 43.0 & \bf 40.6 & 41.4 & 43.6 \\
LaBSE & \bf 81.0 & \bf 81.4 & 81.4 & 74.8 & 75.6 & 76.0 & 42.6 & 42.5 & 43.4 & 37.8 & 41.7 & 43.3 \\
LASER2 & 80.8 & 81.3 & \bf 81.5 & 74.4 & \bf 76.3 & \bf 76.2 & 39.6 & 40.2 & 41.3 & 35.3 & 40.1 & 42.5 \\
SONAR & 80.8 & 81.3 & 81.3 & 74.9 & 75.9 & 76.0 & \bf 43.4 & \bf 43.4 & \bf 44.4 & 39.7 & 41.6 & \bf 44.5 \\
Random & 80.8 & 80.8 & 80.7 & 74.3 & 75.3 & 75.4 & 36.2 & 36.8 & 37.1 & 30.1 & 33.7 & 37.0 \\
\midrule
LLaMA~2~13B \\
\midrule
Embed v3 & \bf 87.2 & 87.3 & \bf 87.6 & 85.9 & 85.9 & 86.3 & 43.4 & 46.3 & \bf 47.8 & 40.9 & 42.6 & 44.1 \\
E5 & 87.1 & 87.3 & 87.5 & \bf 86.0 & 86.2 & 86.4 & \bf 43.5 & 46.1 & 47.6 & \bf 41.7 & 43.4 & 43.5 \\
LaBSE & \bf 87.2 & \bf 87.4 & 87.4 & 85.7 & 86.2 & \bf 86.6 & 42.5 & 45.6 & 47.4 & 39.2 & 42.2 & 43.4 \\
LASER2 & 87.0 & 87.2 & 87.4 & 85.7 & \bf 86.3 & 86.2 & 41.7 & 43.7 & 45.2 & 36.7 & 41.7 & 42.8 \\
SONAR & 87.2 & 87.1 & 87.4 & 85.7 & 86.0 & \bf 86.6 & 43.0 & \bf 46.4 & 47.7 & 39.6 & \bf 44.9 & \bf 44.5 \\
Random & 86.9 & 87.2 & 87.4 & 85.7 & 85.9 & 86.2 & 38.8 & 39.9 & 40.7 & 29.4 & 34.8 & 36.3 \\
\midrule
LLaMA~2~70B \\
\midrule
Embed v3 & 87.5 & 88.0 & 88.1 & \bf 87.2 & 87.6 & 87.7 & 53.5 & \bf 61.2 & 62.6 & 41.1 & 47.7 & \bf 49.4 \\
E5 & \bf 87.7 & 88.1 & \bf 88.3 & \bf 87.2 & 87.5 & \bf 87.8 & 53.0 & 61.0 & \bf 62.9 & \bf 41.7 & \bf 48.4 & 48.4 \\
LaBSE & 87.5 & \bf 88.2 & 88.2 & 87.0 & 87.7 & 87.7 & \bf 53.6 & 60.6 & 62.3 & 40.1 & 48.2 & 48.4 \\
LASER2 & 87.5 & 88.0 & 88.2 & \bf 87.2 & 87.6 & 87.7 & 53.0 & 59.5 & 60.8 & 39.1 & 46.9 & 47.7 \\
SONAR & \bf 87.7 & 88.1 & \bf 88.3 & \bf 87.2 & \bf 87.8 & 87.7 & 52.6 & 60.8 & 62.6 & 41.3 & 48.1 & 48.9 \\
Random & 87.4 & 87.9 & 88.1 & 87.1 & 87.4 & 87.6 & 49.4 & 56.5 & 57.3 & 34.2 & 40.0 & 41.9 \\
\midrule
Mixtral~8x7B~v0.1 \\
\midrule
Embed v3 & \bf 88.2 & \bf 88.4 & \bf 88.5 & 87.6 & 87.9 & 88.1 & 53.3 & 56.9 & \bf 59.8 & 34.1 & 45.2 & 47.9 \\
E5 & 88.0 & \bf 88.4 & 88.4 & 87.5 & \bf 88.2 & \bf 88.3 & \bf 53.5 & 56.5 & 59.4 & \bf 34.3 & \bf 45.3 & 47.5 \\
LaBSE & 88.2 & \bf 88.4 & \bf 88.5 & \bf 87.8 & 88.1 & 88.1 & 53.1 & 56.8 & 58.8 & 32.9 & \bf 45.3 & 47.7 \\
LASER2 & 88.0 & 88.3 & 88.4 & 87.5 & \bf 88.2 & 88.0 & 51.5 & 55.5 & 57.6 & 32.4 & 44.5 & 47.2 \\
SONAR & \bf 88.2 & \bf 88.4 & \bf 88.5 & 87.2 & 88.0 & \bf 88.3 & 53.3 & \bf 57.0 & 58.7 & 33.3 & 45.1 & \bf 48.1 \\
Random & 88.0 & 88.2 & 88.3 & 87.4 & 88.0 & 88.1 & 50.3 & 52.2 & 53.5 & 25.6 & 37.9 & 40.9 \\
\bottomrule
\end{tabular}
\end{center}
\caption{Additional results (other LLMs): laCOMET results for example retrieval with different sentence embeddings in $k$-shot settings ($k \in \{1, 5, 10\}$).}
\label{tab:more-lacomet-se}
\end{table*}

\begin{table*}[ht]
\centering\small
\begin{tabular}{lcccccccccccc}
\toprule
\multicolumn{1}{c}{\bf }  & \multicolumn{3}{c}{\bf eng$\rightarrow$fra} & \multicolumn{3}{c}{\bf eng$\rightarrow$deu} & \multicolumn{3}{c}{\bf eng$\rightarrow$swh} & \multicolumn{3}{c}{\bf eng$\rightarrow$wol}\\
 {} & 1 & 5 & 10 & 1 & 5 & 10 & 1 & 5 & 10 & 1 & 5 & 10\\
\midrule
Gemma~2B \\
\midrule
SONAR & \bf 84.8 & \bf 85.2 & \bf 85.3 & \bf 82.0 & 83.2 & \bf 83.5 & 63.7 & \bf 67.3 & \bf 68.5 & \bf 38.2 & \bf 44.5 & \bf 47.4 \\
BM25 & 84.7 & 85.1 & 85.2 & 81.9 & 83.0 & 83.1 & \bf 64.1 & \bf 67.3 & 68.4 & 36.3 & 43.4 & 45.3 \\
R-BM25 & 84.5 & 84.9 & 84.8 & \bf 82.0 & 83.0 & 83.1 & 63.1 & 64.5 & 65.1 & 24.4 & 33.3 & 35.9 \\
BLEU & 84.7 & 85.0 & 85.1 & 81.8 & 83.2 & 82.9 & 63.6 & 67.0 & 67.0 & 34.1 & 42.1 & 43.3 \\
RoBERTa & \bf 84.8 & 85.0 & 85.0 & 81.8 & \bf 83.3 & \bf 83.5 & 63.3 & 65.8 & 66.0 & 31.9 & 40.3 & 43.3 \\
Random & 84.6 & 84.7 & 84.9 & 81.7 & 82.7 & 83.0 & 62.3 & 64.4 & 65.1 & 26.8 & 35.2 & 37.7 \\
\midrule
OLMo~7B \\
\midrule
SONAR & 80.8 & 81.3 & \bf 81.3 & \bf 74.9 & \bf 75.9 & \bf 76.0 & \bf 43.4 & \bf 43.4 & \bf 44.4 & \bf 39.7 & \bf 41.6 & \bf 44.5 \\
BM25 & 80.7 & \bf 81.4 & 81.1 & 74.6 & 75.5 & 75.7 & 40.4 & 41.1 & 42.1 & 36.9 & 40.3 & 42.3 \\
R-BM25 & 80.2 & 80.7 & 80.8 & 74.3 & 75.1 & 75.1 & 35.6 & 36.6 & 37.0 & 24.6 & 30.3 & 34.7 \\
BLEU & \bf 80.9 & 81.1 & 81.0 & \bf 74.9 & 75.3 & 75.8 & 39.8 & 40.5 & 41.1 & 35.5 & 40.2 & 42.4 \\
RoBERTa & 80.8 & 81.0 & 80.9 & 74.4 & 75.6 & 75.2 & 39.7 & 38.6 & 39.3 & 35.8 & 37.9 & 39.9 \\
Random & 80.8 & 80.8 & 80.7 & 74.3 & 75.3 & 75.4 & 36.2 & 36.8 & 37.1 & 30.1 & 33.7 & 37.0 \\
\midrule
LLaMA~2~13B \\
\midrule
SONAR & \bf 87.2 & 87.1 & 87.4 & 85.7 & 86.0 & \bf 86.6 & \bf 43.0 & \bf 46.4 & \bf 47.7 & \bf 39.6 & \bf 44.9 & \bf 44.5 \\
BM25 & 86.9 & 87.0 & 87.3 & \bf 86.1 & 85.8 & 86.5 & 41.2 & 44.9 & 46.4 & 38.3 & 41.5 & 43.4 \\
R-BM25 & 87.1 & \bf 87.2 & 87.3 & 85.9 & \bf 86.1 & 86.4 & 38.5 & 38.9 & 40.1 & 27.3 & 33.2 & 34.6 \\
BLEU & 87.0 & 86.4 & 87.3 & 85.7 & 85.5 & 86.5 & 40.8 & 44.0 & 44.6 & 36.0 & 41.9 & 42.7 \\
RoBERTa & 87.0 & 87.0 & \bf 87.5 & 85.9 & 85.7 & 86.3 & 40.6 & 42.2 & 43.2 & 36.9 & 39.7 & 40.3 \\
Random & 86.9 & \bf 87.2 & 87.4 & 85.7 & 85.9 & 86.2 & 38.8 & 39.9 & 40.7 & 29.4 & 34.8 & 36.3 \\
\midrule
LLaMA~2~70B \\
\midrule
SONAR & \bf 87.7 & \bf 88.1 & \bf 88.3 & \bf 87.2 & \bf 87.8 & \bf 87.7 & \bf 52.6 & \bf 60.8 & \bf 62.6 & \bf 41.3 & \bf 48.1 & \bf 48.9 \\
BM25 & \bf 87.7 & 87.9 & 88.1 & 86.9 & 87.6 & 87.7 & 50.6 & 60.1 & 61.8 & 37.9 & 45.8 & 48.7 \\
R-BM25 & 87.3 & 87.8 & 88.0 & 87.1 & 87.5 & 87.6 & 47.0 & 56.1 & 57.7 & 29.9 & 38.4 & 41.3 \\
BLEU & 87.2 & 88.0 & 88.1 & \bf 87.2 & 87.4 & 87.6 & 50.7 & 59.4 & 60.1 & 38.7 & 45.8 & 46.5 \\
RoBERTa & 87.4 & 87.9 & 88.2 & 87.1 & 87.5 & 87.5 & 51.8 & 58.0 & 59.0 & 39.4 & 44.8 & 45.8 \\
Random & 87.4 & 87.9 & 88.1 & 87.1 & 87.4 & 87.6 & 49.4 & 56.5 & 57.3 & 34.2 & 40.0 & 41.9 \\
\midrule
Mixtral~8x7B~v0.1 \\
\midrule
SONAR & \bf 88.2 & 88.4 & \bf 88.5 & 87.2 & \bf 88.0 & \bf 88.3 & \bf 53.3 & \bf 57.0 & \bf 58.7 & \bf 33.3 & \bf 45.1 & \bf 48.1 \\
BM25 & 87.9 & 88.3 & 88.2 & \bf 87.6 & \bf 88.0 & 88.1 & 52.4 & 56.9 & 58.0 & 30.6 & 44.9 & 46.8 \\
R-BM25 & 88.0 & 88.2 & 88.3 & 87.4 & 87.9 & 88.0 & 50.2 & 52.9 & 53.6 & 22.8 & 34.5 & 37.5 \\
BLEU & 87.8 & 88.4 & 88.4 & 87.4 & \bf 88.0 & 88.1 & 51.4 & 55.9 & 56.9 & 30.6 & 43.4 & 46.3 \\
RoBERTa & 88.1 & \bf 88.5 & \bf 88.5 & 87.4 & 87.9 & 88.0 & 51.9 & 54.4 & 55.2 & 31.1 & 41.7 & 45.2 \\
Random & 88.0 & 88.2 & 88.3 & 87.4 & \bf 88.0 & 88.1 & 50.3 & 52.2 & 53.5 & 25.6 & 37.9 & 40.9 \\
\bottomrule
\end{tabular}
\caption{Comparison of $k$-shot ($k \in \{1, 5, 10\}$) example retrieval with SONAR to baseline methods (laCOMET).}
\label{tab:more-lacomet-comparison}
\end{table*}

\subsection{Source-to-target example retrieval}\label{appendix:section-s2t}

As mentioned in the main text of the article, we mainly explored source-to-source retrieval (comparing the source sentence to the source side of pool examples). In this section, we provide results for source-to-target retrieval. Tables~\ref{tab:lacomet-se-s2t} and~\ref{tab:more-lacomet-se-s2t} summarize the laCOMET scores obtained using different sentence embeddings with nine LLMs. Example retrieval via similarity search outperforms random sampling, with most gains observed when translating into Swahili or Wolof. SONAR does even better in this setup and we attribute this to its cross-lingual training which covers all the languages we experiment with. Comparing example retrieval in source-to-source and source-to-target does not allow us to draw systematic conclusions. However, the performance of both approaches are similar when translating into high-resource languages. When translating into low-resource languages, some sentence embeddings tend (e.g. LaBSE) to perform worse for source-to-target than for source-to-source, which is typically related to the amount of data in the language seen during training.

\begin{table*}[ht]
\centering\small
\begin{center}
\begin{tabular}{lcccccccccccc}
\toprule
\multicolumn{1}{c}{\bf }  & \multicolumn{3}{c}{\bf eng$\rightarrow$fra} & \multicolumn{3}{c}{\bf eng$\rightarrow$deu} & \multicolumn{3}{c}{\bf eng$\rightarrow$swh} & \multicolumn{3}{c}{\bf eng$\rightarrow$wol}\\
 {} & 1 & 5 & 10 & 1 & 5 & 10 & 1 & 5 & 10 & 1 & 5 & 10\\
\midrule
BLOOM~7B1 \\
\midrule
Embed v3 & 79.9 & 86.7 & \bf 86.8 & 55.7 & \bf 60.4 & 60.9 & 58.0 & 68.3 & 68.9 & 48.4 & 50.0 & 50.6 \\
E5 & \bf 80.0 & 86.5 & 86.6 & 54.7 & 59.9 & 60.5 & 58.8 & 67.6 & 69.0 & 47.8 & 49.0 & 49.9 \\
LaBSE & 79.2 & 86.6 & 86.6 & 54.9 & 60.1 & 60.5 & 57.9 & \bf 68.5 & \bf 69.4 & 46.4 & 47.4 & 48.6 \\
LASER2 & 78.7 & \bf 86.9 & 86.7 & 54.6 & 60.0 & 59.9 & \bf 58.9 & 67.7 & 68.3 & 50.4 & 50.8 & 51.0 \\
SONAR & 79.8 & 86.6 & 86.6 & \bf 55.9 & 60.1 & \bf 61.5 & 57.8 & 68.1 & 68.9 & \bf 50.9 & \bf 51.2 & \bf 52.1 \\
Random & 77.3 & 86.5 & 86.6 & 52.8 & 57.7 & 57.7 & 56.9 & 65.1 & 66.0 & 46.5 & 45.1 & 46.4 \\
\midrule
Mistral~7B~v0.1\\
\midrule
Embed v3 & 85.9 & \bf 87.0 & \bf 87.0 & 83.2 & 85.7 & 85.7 & 37.0 & \bf 41.3 & \bf 44.3 & 34.4 & 41.8 & 43.2 \\
E5 & 86.0 & 86.5 & \bf 87.0 & 82.7 & 85.4 & 85.7 & 36.8 & 40.9 & 42.0 & 34.4 & 40.8 & 43.9 \\
LaBSE & \bf 86.2 & \bf 87.0 & 86.9 & \bf 83.9 & 85.3 & 85.7 & 37.2 & 39.6 & 42.8 & 28.0 & 37.6 & 40.3 \\
LASER2 & \bf 86.2 & 86.8 & 86.9 & 83.7 & 85.7 & 85.7 & 34.7 & 37.6 & 39.0 & 32.4 & 41.9 & 42.8 \\
SONAR & 86.1 & 86.8 & \bf 87.0 & 83.6 & \bf 85.8 & \bf 86.0 & \bf 37.4 & 40.9 & 42.8 & \bf 35.3 & \bf 44.1 & \bf 44.5 \\
Random & 85.8 & 86.5 & 86.6 & 83.0 & 85.4 & 85.5 & 32.7 & 33.5 & 33.8 & 26.7 & 33.2 & 36.0 \\
\midrule
LLaMA~2~7B \\
\midrule
Embed v3 & 85.7 & 86.2 & \bf 86.3 & 84.0 & 85.1 & \bf 85.3 & \bf 46.3 & \bf 44.2 & \bf 45.8 & 37.9 & 43.1 & 45.1 \\
E5 & \bf 85.8 & 86.1 & \bf 86.3 & 83.8 & 84.8 & 85.0 & 44.8 & 43.2 & 45.0 & 37.5 & 41.7 & 44.9 \\
LaBSE & 85.5 & 86.2 & \bf 86.3 & \bf 84.1 & 85.0 & \bf 85.3 & 43.7 & 42.6 & 45.2 & 33.7 & 38.5 & 39.2 \\
LASER2 & \bf 85.8 & 86.1 & 86.1 & 83.9 & 85.2 & \bf 85.2 & 40.6 & 38.8 & 40.9 & 41.2 & 43.8 & 45.2 \\
SONAR & 85.7 & \bf 86.3 & \bf 86.3 & 84.0 & 85.1 & 85.2 & 45.8 & 43.2 & 45.4 & \bf 40.8 & \bf 45.1 & \bf 46.3 \\
Random & 85.6 & 85.9 & 86.0 & 83.6 & 84.8 & 85.0 & 35.4 & 34.7 & 35.8 & 34.4 & 34.7 & 36.5 \\
\midrule
Gemma~7B \\
\midrule
Embed v3 & \bf 87.7 & 88.0 & 88.0 & 86.8 & 87.3 & 87.6 & \bf 79.4 & \bf 80.7 & 80.7 & 35.6 & 43.0 & 46.5 \\
E5 & 87.6 & 87.9 & \bf 88.1 & 86.6 & 87.4 & 87.6 & \bf 79.4 & 80.5 & 80.8 & 35.6 & 42.3 & 46.0 \\
LaBSE & 87.6 & \bf 88.1 & 87.9 & 87.0 & \bf 87.6 & 87.6 & 79.1 & 80.4 & \bf 81.0 & 33.5 & 41.7 & 44.7 \\
LASER2 & 87.5 & 88.0 & 88.3 & \bf 87.1 & 87.5 & \bf 87.7 & 79.1 & 79.9 & 80.6 & 33.9 & 42.6 & 46.0 \\
SONAR & 87.6 & 88.0 & \bf 88.1 & 86.7 & 87.5 & \bf 87.7 & \bf 79.4 & 80.3 & 80.7 & \bf 37.0 & \bf 44.1 & \bf 47.7 \\
Random & 87.5 & 87.9 & 88.0 & 86.6 & 87.2 & 87.3 & 78.4 & 79.6 & 79.8 & 30.9 & 37.4 & 40.5 \\
\bottomrule
\end{tabular}
\end{center}
\caption{laCOMET scores of $k$-shot ($k \in \{1, 5, 10\}$) \textit{source-to-target} example retrieval  with different sentence embeddings for 4 LLMs (BLOOM~7B1, Mistral~7B~v0.1, LLaMA~2~7B and Gemma~7B).}
\label{tab:lacomet-se-s2t}
\end{table*}


\begin{table*}[ht]
\centering\small
\begin{center}
\begin{tabular}{lcccccccccccc}
\toprule
\multicolumn{1}{c}{\bf }  & \multicolumn{3}{c}{\bf eng $\rightarrow$ fra} & \multicolumn{3}{c}{\bf eng $\rightarrow$ deu} & \multicolumn{3}{c}{\bf eng $\rightarrow$ swh} & \multicolumn{3}{c}{\bf eng $\rightarrow$ wol}\\
 {} & 1 & 5 & 10 & 1 & 5 & 10 & 1 & 5 & 10 & 1 & 5 & 10\\
\midrule
Gemma~2B \\
\midrule
Embed v3 & 84.8 & \bf 85.1 & \bf 85.4 & \bf 82.2 & 83.2 & 83.2 & \bf 64.1 & 66.7 & \bf 68.6 & 34.3 & \bf 43.0 & 45.4 \\
E5 & \bf 84.9 & \bf 85.1 & 85.3 & 81.8 & 82.8 & 83.1 & 63.8 & 66.8 & 68.2 & \bf 34.5 & 42.3 & \bf 45.6 \\
LaBSE & 84.7 & \bf 85.1 & 85.2 & 82.1 & \bf 83.4 & \bf 83.6 & 63.7 & \bf 67.3 & 67.8 & 29.5 & 38.9 & 41.1 \\
LASER2 & 84.7 & \bf 85.1 & 85.2 & \bf 82.2 & 83.2 & 83.4 & 63.2 & 66.1 & 66.5 & 33.8 & 42.6 & 44.8 \\
SONAR & 84.7 & \bf 85.1 & 85.2 & \bf 82.2 & 83.2 & 83.4 & 63.2 & 66.1 & 66.5 & 33.8 & 42.6 & 44.8 \\
Random & 84.6 & 84.7 & 84.9 & 81.7 & 82.7 & 83.0 & 62.3 & 64.4 & 65.1 & 26.8 & 35.2 & 37.7 \\
\midrule
OLMo~7B \\
\midrule
Embed v3 & \bf 81.0 & 81.3 & \bf 81.3 & 74.7 & 75.6 & 75.7 & 43.0 & \bf 43.3 & \bf 44.2 & 37.0 & 40.3 & 42.4 \\
E5 & 80.9 & \bf 81.5 & \bf 81.3 & 74.5 & 75.5 & 75.4 & 42.4 & 42.5 & 43.7 & 37.4 & 38.5 & 41.1 \\
LaBSE & 80.8 & 81.3 & 81.2 & 74.8 & \bf 76.0 & \bf 76.0 & 41.8 & 42.1 & 43.8 & 31.7 & 37.8 & 40.5 \\
LASER2 & 80.8 & 81.4 & 81.1 & \bf 74.9 & 75.8 & 75.9 & 39.6 & 39.9 & 41.0 & 35.1 & 39.5 & 41.0 \\
SONAR & \bf 81.0 & 81.1 & 81.0 & \bf 74.9 & 75.7 & 75.8 & \bf 43.8 & 42.9 & 43.9 & \bf 38.9 & \bf 42.2 & \bf 43.1 \\
Random & 80.8 & 80.8 & 80.7 & 74.3 & 75.3 & 75.4 & 36.2 & 36.8 & 37.1 & 30.1 & 33.7 & 37.0 \\
\midrule
LLaMA~2~13B \\
\midrule
Embed v3 & \bf 87.2 & 87.2 & \bf 87.6 & 85.9 & 86.1 & 86.5 & 42.1 & 46.1 & 47.4 & 37.9 & 42.2 & 42.6 \\
E5 & 87.1 & 86.9 & 87.3 & 85.6 & 86.1 & 86.3 & 42.3 & 45.8 & 47.2 & 37.4 & 40.9 & 42.0 \\
LaBSE & 87.1 & \bf 87.4 & 87.5 & \bf 86.1 & \bf 86.3 & \bf 86.6 & 42.8 & 45.8 & 47.7 & 32.7 & 40.6 & 40.6 \\
LASER2 & 87.1 & 87.3 & 87.5 & 85.9 & 86.2 & 86.4 & 40.3 & 43.1 & 43.9 & 35.0 & 40.4 & 41.2 \\
SONAR & \bf 87.2 & 87.0 & 87.5 & 85.7 & 86.0 & 86.4 & \bf 43.0 & \bf 46.6 & \bf 48.4 & \bf 39.7 & \bf 43.9 & \bf 44.8 \\
Random & 86.9 & 87.2 & 87.4 & 85.7 & 85.9 & 86.2 & 38.8 & 39.9 & 40.7 & 29.4 & 34.8 & 36.3 \\
\midrule
LLaMA~2~70B \\
\midrule
Embed v3 & 87.6 & 88.1 & 88.2 & 87.1 & 87.3 & \bf 87.8 & 53.3 & 61.0 & 62.3 & 38.9 & 46.7 & 47.5 \\
E5 & 87.7 & 88.0 & 88.2 & 87.0 & 87.5 & 87.6 & 52.0 & 60.5 & 62.4 & 38.6 & 45.7 & 47.7 \\
LaBSE & \bf 87.8 & \bf 88.2 & 88.2 & 87.3 & 87.5 & 87.6 & \bf 53.5 & 60.3 & 62.3 & 37.2 & 44.0 & 46.0 \\
LASER2 & 87.5 & \bf 88.2 & 88.2 & \bf 87.4 & \bf 87.7 & \bf 87.8 & 51.2 & 59.0 & 60.1 & 40.0 & 46.1 & 46.5 \\
SONAR & 87.7 & \bf 88.2 & \bf 88.3 & 87.2 & 87.5 & 87.6 & 52.5 & \bf 61.6 & \bf 62.9 & \bf 41.7 & \bf 48.2 & \bf 49.5 \\
Random & 87.4 & 87.9 & 88.1 & 87.1 & 87.4 & 87.6 & 49.4 & 56.5 & 57.3 & 34.2 & 40.0 & 41.9 \\
\midrule
Mixtral~8x7B~v0.1 \\
\midrule
Embed v3 & \bf 88.3 & 88.4 & 88.4 & 87.4 & \bf 88.1 & \bf 88.3 & \bf 53.4 & 57.1 & \bf 59.4 & 31.7 & 45.1 & 47.2 \\
E5 & 88.2 & 88.4 & 88.3 & 87.3 & \bf 88.1 & 88.1 & 52.2 & 56.3 & 59.0 & 29.6 & 43.2 & 45.6 \\
LaBSE & \bf 88.3 & 88.4 & 88.4 & \bf 87.7 & \bf 88.1 & 88.1 & 53.3 & 56.1 & 58.7 & 28.3 & 41.1 & 44.5 \\
LASER2 & 87.9 & 88.4 & \bf 88.6 & 87.6 & \bf 88.1 & 88.1 & 51.6 & 55.1 & 56.3 & 30.6 & 43.6 & 45.4 \\
SONAR & 88.2 & \bf 88.5 & \bf 88.6 & 87.6 & 88.0 & 88.2 & 53.3 & \bf 57.3 & \bf 59.4 & \bf 34.5 & \bf 46.1 & \bf 47.9 \\
Random & 88.0 & 88.2 & 88.3 & 87.4 & 88.0 & 88.1 & 50.3 & 52.2 & 53.5 & 25.6 & 37.9 & 40.9 \\
\bottomrule
\end{tabular}
\end{center}
\caption{Benchmarking of example retrieval \textit{source-to-target} with different sentence embeddings in $k$-shot ($k \in \{1, 5, 10\}$). We report the laCOMET scores.}
\label{tab:more-lacomet-se-s2t}
\end{table*}

\subsection{Translation into English} \label{appendix:to-english}

In this section, we benchmark example retrieval with different sentence embeddings for fra$\rightarrow$eng, deu$\rightarrow$eng, swh$\rightarrow$eng and wol$\rightarrow$eng. Tables~\ref{tab:to-english-bleu},~\ref{tab:to-english-comet} and~\ref{tab:to-english-lacomet} respectively contain the BLEU, COMET and laCOMET scores obtained with BLOOM~7B1 and LLaMA~2~7B. In this scenario, example retrieval via similarity search also proves beneficial, especially when the source language is a mid- or low-resource language. The gains are significant, but not as highly as for the opposite translation direction. In summary, the conclusions are generally consistent with those for the opposite direction.

\begin{table*}[ht]
\centering\small
\begin{center}
\begin{tabular}{lcccccccccccc}
\toprule
\multicolumn{1}{c}{\bf }  & \multicolumn{3}{c}{\bf fra $\rightarrow$ eng} & \multicolumn{3}{c}{\bf deu $\rightarrow$ eng} & \multicolumn{3}{c}{\bf swh $\rightarrow$ eng} & \multicolumn{3}{c}{\bf wol $\rightarrow$ eng}\\
 {} & 1 & 5 & 10 & 1 & 5 & 10 & 1 & 5 & 10 & 1 & 5 & 10\\
\midrule
BLOOM~7B1 \\
\midrule
Embed v3 & 44.3 & 45.2 & 45.0 & 31.2 & 31.8 & \bf 32.4 & 27.7 & 28.8 & \bf 28.7 & 5.6 & 6.8 & 6.8 \\
E5 & \bf 44.4 & \bf 45.3 & \bf 45.5 & \bf 31.5 & 31.9 & \bf 32.4 & 27.6 & 28.8 & 28.5 & 5.6 & 7.1 & 6.6 \\
LaBSE & 44.1 & \bf 45.3 & 45.2 & 31.1 & 32.0 & 32.1 & 27.7 & \bf 29.0 & 28.4 & 5.7 & 6.8 & 6.4 \\
LASER2 & 44.2 & 44.8 & 44.6 & 31.2 & 31.4 & 31.8 & \bf 27.8 & 28.3 & 28.3 & 5.3 & 6.8 & 6.6 \\
SONAR & 44.3 & 45.2 & 45.1 & 31.2 & \bf 32.3 & 32.3 & 27.4 & 28.7 & 28.6 & \bf 6.2 & \bf 7.2 & \bf 7.2 \\
Random & 44.0 & 45.1 & 45.0 & 30.6 & 31.2 & 31.1 & 27.6 & 28.5 & 28.4 & 5.4 & 6.7 & 6.6 \\
\midrule
LLaMA~2~7B \\
\midrule
Embed v3 & 44.9 & 46.4 & 46.8 & 43.7 & 45.0 & 45.6 & 9.2 & 10.9 & 11.3 & 6.1 & 7.1 & 7.2 \\
E5 & 45.1 & 46.5 & 47.0 & 43.8 & 45.5 & \bf 45.7 & \bf 9.4 & 11.0 & 11.3 & 6.4 & 7.3 & 7.4 \\
LaBSE & 45.3 & 46.7 & 47.2 & \bf 43.9 & 45.0 & 45.5 & 9.2 & \bf 11.2 & \bf 11.4 & 6.3 & 7.2 & 7.4 \\
LASER2 & 45.0 & \bf 46.9 & 47.1 & 43.7 & 45.3 & 45.4 & 8.7 & 10.2 & 10.5 & \bf 6.7 & 7.4 & \bf 7.6 \\
SONAR & \bf 45.4 & 46.8 & \bf 47.3 & 43.4 & \bf 45.5 & 45.6 & 9.2 & 10.9 & \bf 11.4 & \bf 6.7 & \bf 7.5 & 7.4 \\
Random & 44.5 & 45.9 & 46.6 & 43.6 & 45.1 & 45.2 & 8.7 & 9.7 & 9.8 & 6.0 & 7.0 & 6.9 \\
\bottomrule
\end{tabular}
\end{center}
\caption{Benchmarking of example retrieval with different sentence embeddings in $k$-shot ($k \in \{1, 5, 10\}$). We report the BLEU scores.}
\label{tab:to-english-bleu}
\end{table*}

\begin{table*}[ht]
\centering\small
\begin{center}
\begin{tabular}{lcccccccccccc}
\toprule
\multicolumn{1}{c}{\bf }  & \multicolumn{3}{c}{\bf fra $\rightarrow$ eng} & \multicolumn{3}{c}{\bf deu $\rightarrow$ eng} & \multicolumn{3}{c}{\bf swh $\rightarrow$ eng} & \multicolumn{3}{c}{\bf wol $\rightarrow$ eng}\\
 {} & 1 & 5 & 10 & 1 & 5 & 10 & 1 & 5 & 10 & 1 & 5 & 10\\
\midrule
BLOOM~7B1 \\
\midrule
Embed v3 & 88.2 & \bf 88.4 & 88.4 & \bf 82.2 & 82.9 & \bf 83.4 & 77.5 & 78.7 & 79.2 & 48.8 & 50.8 & 51.4 \\
E5 & 88.1 & \bf 88.4 & 88.4 & \bf 82.2 & 82.8 & 83.3 & 77.4 & 79.0 & 79.2 & 48.5 & 50.7 & 51.1 \\
LaBSE & \bf 88.3 & \bf 88.4 & 88.4 & 81.6 & 82.5 & 82.7 & 77.4 & 78.7 & 78.9 & 47.0 & 49.1 & 49.3 \\
LASER2 & \bf 88.3 & 88.3 & 88.3 & 81.6 & 82.1 & 82.3 & 77.4 & 78.4 & 78.7 & 47.3 & 49.6 & 49.9 \\
SONAR & 88.2 & 88.3 & \bf 88.5 & 82.0 & \bf 83.0 & 83.2 & \bf 77.7 & \bf 79.1 & \bf 79.6 & \bf 49.2 & \bf 51.3 & \bf 51.7 \\
Random & 88.2 & \bf 88.4 & 88.3 & 81.1 & 81.7 & 81.7 & 77.1 & 78.1 & 78.4 & 45.1 & 47.4 & 47.9 \\
\midrule
LLaMA~2~7B \\
\midrule
Embed v3 & 88.6 & 88.9 & 89.0 & \bf 88.5 & \bf 88.8 & 88.8 & \bf 59.8 & \bf 63.4 & \bf 64.2 & 48.8 & 50.4 & 51.4 \\
E5 & 88.6 & 88.9 & 89.0 & \bf 88.5 & 88.7 & \bf 88.9 & 59.4 & 62.9 & 63.7 & 48.7 & 50.8 & 51.6 \\
LaBSE & \bf 88.7 & 88.9 & 89.0 & \bf 88.5 & \bf 88.8 & 88.8 & 59.0 & 62.7 & 63.1 & 47.2 & 49.1 & 50.0 \\
LASER2 & \bf 88.7 & 88.9 & 89.0 & 88.4 & \bf 88.8 & 88.8 & 57.7 & 60.3 & 61.1 & 47.6 & 49.7 & 50.3 \\
SONAR & \bf 88.7 & \bf 89.0 & \bf 89.1 & \bf 88.5 & \bf 88.8 & 88.8 & 59.7 & 63.3 & \bf 64.2 & \bf 49.2 & \bf 51.5 & \bf 51.9 \\
Random & 88.6 & 88.8 & 88.9 & 88.4 & 88.7 & 88.7 & 56.1 & 58.0 & 58.8 & 45.2 & 47.6 & 48.2 \\
\bottomrule
\end{tabular}
\end{center}
\caption{COMET scores for $k$-shot ($k \in \{1, 5, 10\}$) example retrieval with different sentence embeddings.}
\label{tab:to-english-comet}
\end{table*}

\begin{table*}[ht]
\centering\small
\begin{center}
\begin{tabular}{lcccccccccccc}
\toprule
\multicolumn{1}{c}{\bf }  & \multicolumn{3}{c}{\bf fra $\rightarrow$ eng} & \multicolumn{3}{c}{\bf deu $\rightarrow$ eng} & \multicolumn{3}{c}{\bf swh $\rightarrow$ eng} & \multicolumn{3}{c}{\bf wol $\rightarrow$ eng}\\
 {} & 1 & 5 & 10 & 1 & 5 & 10 & 1 & 5 & 10 & 1 & 5 & 10\\
\midrule
BLOOM~7B1 \\
\midrule
Embed v3 & 88.2 & 88.4 & 88.4 & 82.1 & 82.9 & \bf 83.4 & \bf 77.3 & 78.6 & 79.0 & 48.6 & 50.6 & 51.2 \\
E5 & 88.1 & 88.4 & 88.4 & \bf 82.2 & 82.8 & 83.3 & 77.2 & 78.8 & 79.0 & 48.3 & 50.5 & 51.0 \\
LaBSE & 88.3 & \bf 88.4 & 88.4 & 81.6 & 82.5 & 82.7 & 76.8 & 78.6 & 78.6 & 46.4 & 48.6 & 49.1 \\
LASER2 & \bf 88.3 & 88.3 & 88.3 & 81.5 & 82.1 & 82.3 & 76.7 & 78.1 & 78.6 & 46.8 & 49.0 & 49.5 \\
SONAR & 88.2 & 88.3 & \bf 88.4 & 82.0 & \bf 83.0 & 83.2 & 77.1 & \bf 78.9 & \bf 79.4 & \bf 48.7 & \bf 51.1 & \bf 51.6 \\
Random & 88.1 & 88.4 & 88.3 & 81.1 & 81.7 & 81.7 & 76.2 & 77.8 & 78.3 & 43.9 & 46.4 & 47.2 \\
\midrule
LLaMA~2~7B \\
\midrule
Embed v3 & 88.6 & 88.9 & 89.0 & \bf 88.5 & \bf 88.8 & 88.8 & \bf 59.3 & \bf 63.4 & \bf 64.1 & \bf 48.5 & 50.2 & 51.3 \\
E5 & 88.6 & 88.9 & 89.0 & \bf 88.5 & 88.7 & \bf 88.9 & 59.1 & 62.8 & 63.7 & 47.9 & 50.7 & \bf 51.6 \\
LaBSE & \bf 88.7 & 88.9 & 89.0 & \bf 88.5 & \bf 88.8 & 88.8 & 58.4 & 62.6 & 63.1 & 46.6 & 48.9 & 49.8 \\
LASER2 & \bf 88.7 & 88.9 & 89.0 & 88.4 & \bf 88.8 & 88.8 & 57.1 & 60.1 & 61.0 & 46.9 & 49.4 & 49.9 \\
SONAR & \bf 88.7 & \bf 89.0 & \bf 89.1 & \bf 88.5 & \bf 88.8 & 88.8 & 59.2 & 63.2 & 64.0 & 48.4 & \bf 51.5 & 51.5 \\
Random & 88.6 & 88.8 & 88.9 & 88.4 & 88.7 & 88.7 & 55.6 & 57.8 & 58.6 & 44.1 & 46.9 & 47.5 \\
\bottomrule
\end{tabular}
\end{center}
\caption{laCOMET scores for $k$-shot ($k \in \{1, 5, 10\}$) example retrieval with different sentence embeddings for into-English language directions.}
\label{tab:to-english-lacomet}
\end{table*}

\subsection{Distribution of issues in zero-shot and few-shot MT} \label{appendix:distribution}

\begin{figure*}[ht]
     \centering
     \begin{subfigure}[b]{\textwidth}
         \centering
         \includegraphics[width=\textwidth]{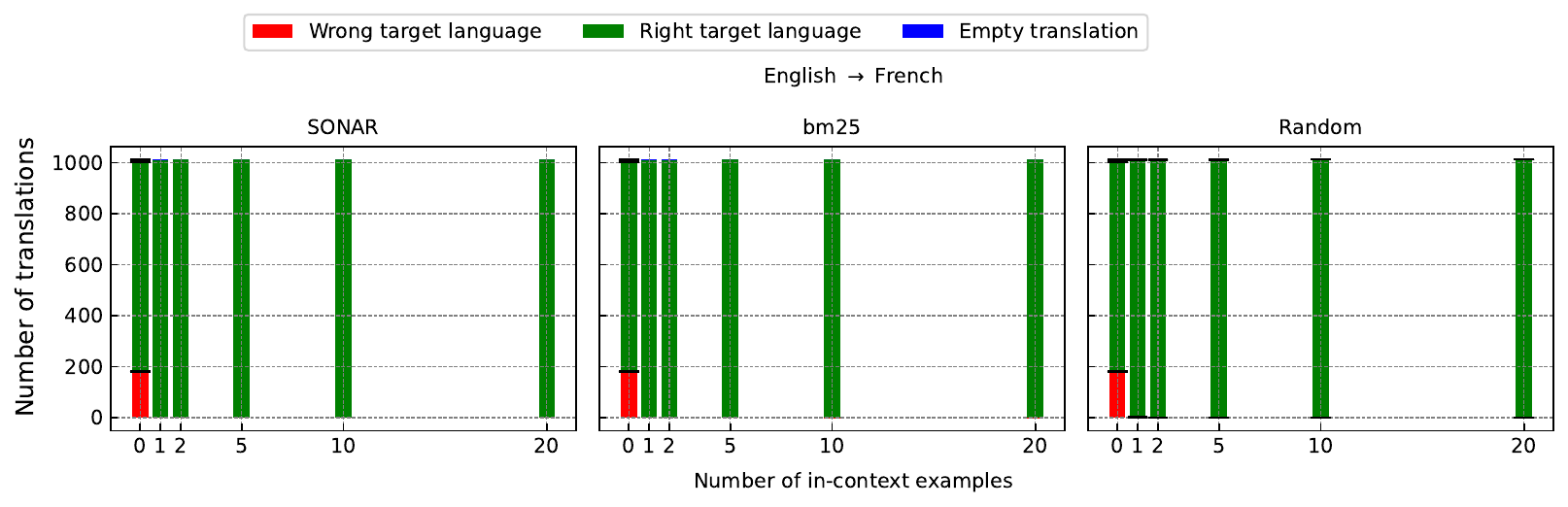}
     \end{subfigure}
     \hfill
     \begin{subfigure}[b]{\textwidth}
         \centering
         \includegraphics[width=\textwidth]{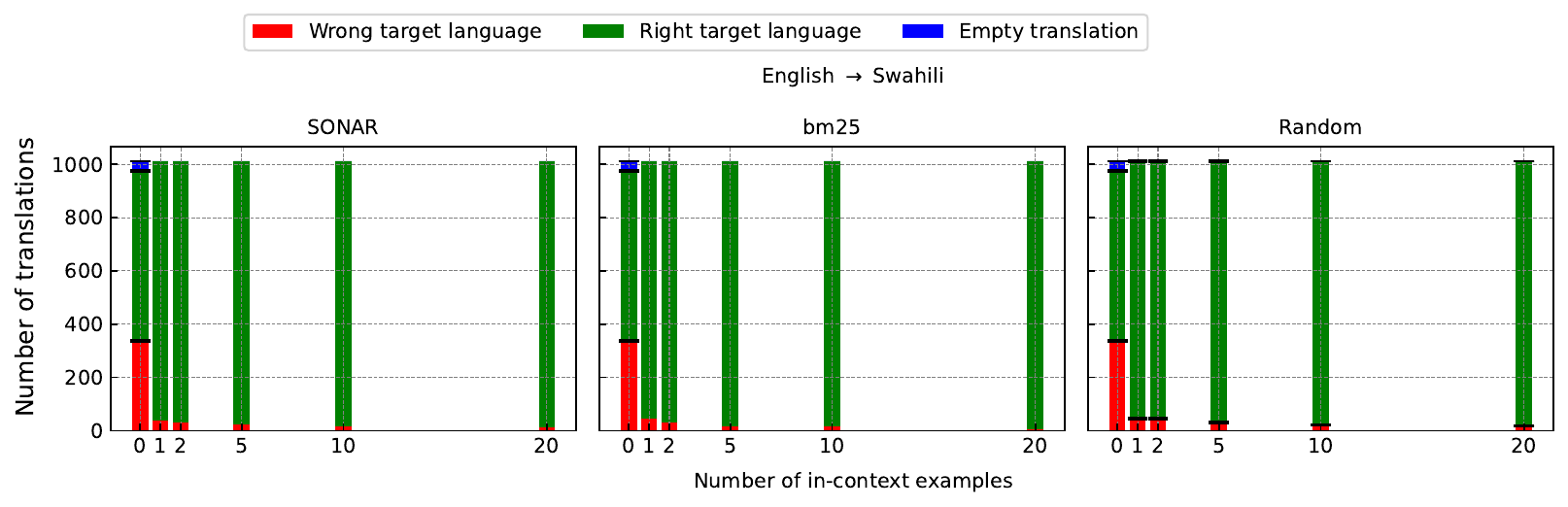}
         \label{fig:lacomet}
     \end{subfigure}
     \hfill
     \begin{subfigure}[b]{\textwidth}
         \centering
         \includegraphics[width=\textwidth]{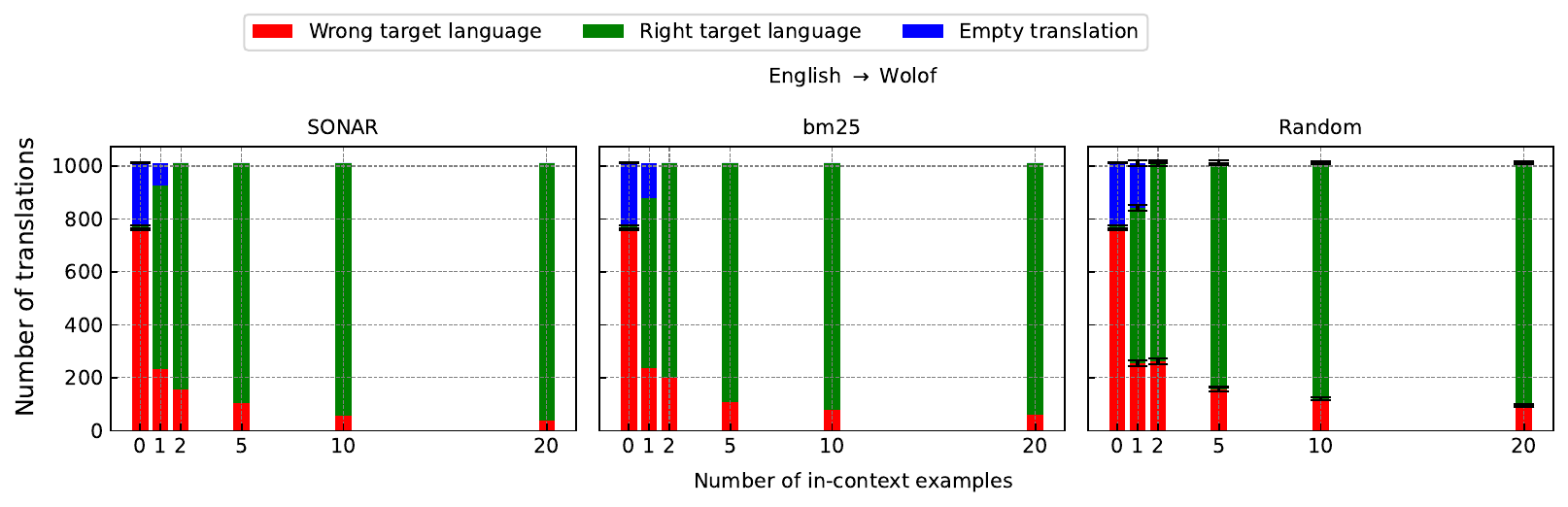}
     \end{subfigure}
     \hfill
        \caption{Error analysis of few-shot translation (eng$\rightarrow$\{fra, swa, wol\}),  of Mixtral~8x7B~v0.1, tracking the number of empty translations, the number of translation in the wrong target language and those in the right language.}
    \label{fig:happens}
\end{figure*}

A major issue when translating with LLMs is the generation of empty translations and translations in the incorrect target language (a problem that appears to decrease as the number of in-context demonstrations increases). We use Mixtral~8x7B~v0.1 to translate from English into French, Swahili, and Wolof. As shown in Figure~\ref{fig:happens}, when translating into French, even a single in-context demonstration ensures that the language model generates a non-empty French sentence in all cases, regardless of whether the demonstrations are chosen randomly. However, for translations into Swahili and Wolof, adding in-context examples does not entirely solve the problem of translating in an incorrect language, although the more in-context demonstrations provided, the less the problem occurs. Moreover, using SONAR and BM25 sampling methods reduces the frequency of these problems compared to random sampling.

\end{document}